%% file: main.tex
\documentclass[10pt,twocolumn,letterpaper]{article}
\usepackage{tikz}
\usepackage[pagenumbers]{cvpr} 

\newcommand\copyrighttext{%
  \footnotesize \textcopyright 2024 IEEE. Personal use of this material is permitted.
  Permission from IEEE must be obtained for all other uses, in any current or future
  media, including reprinting/republishing this material for advertising or promotional
  purposes, creating new collective works, for resale or redistribution to servers or
  lists, or reuse of any copyrighted component of this work in other works.}
\newcommand\copyrightnotice{%
\begin{tikzpicture}[remember picture,overlay]
\node[anchor=south,yshift=10pt] at (current page.south) {\fbox{\parbox{\dimexpr\textwidth-\fboxsep-\fboxrule\relax}{\copyrighttext}}};
\end{tikzpicture}%
}

\newcommand{\red}[1]{{\color{red}#1}}

\usepackage{graphicx}
\usepackage{amsmath}
\usepackage{amssymb}
\usepackage{booktabs}
\usepackage{subfloat}
\usepackage{bigstrut}
\usepackage{multirow}
\usepackage{soul}

\definecolor{cvprblue}{rgb}{0.21,0.49,0.74}
\usepackage[pagebackref,breaklinks,colorlinks,citecolor=cvprblue]{hyperref}

\usepackage[capitalize]{cleveref}
\crefname{section}{Sec.}{Secs.}
\Crefname{section}{Section}{Sections}
\Crefname{table}{Table}{Tables}
\crefname{table}{Tab.}{Tabs.}

\def\paperID{189} 
\def\confName{3DV\xspace}
\def\confYear{2024\xspace}

\newcommand{\blue}[1] {\textcolor[rgb]{0.0,0.0,1.0}{{#1}}}

\def\Put(#1,#2)#3{\leavevmode\makebox(0,0){\put(#1,#2){#3}}}

\usepackage{amsthm}

\theoremstyle{definition}

\def\ie{{\it i.e.}}
\def\eg{{\it e.g.}}
\def\etc{{\it etc.}}
\def\etal{{\it et~al.}}

\title{PIVOT-Net: Heterogeneous Point-Voxel-Tree-based Framework for\\ Point Cloud Compression}

\def\namespacing{15pt}

\author{%
   Jiahao~Pang\hspace{\namespacing}Kevin~Bui\thanks{Work done while the author was an intern at InterDigital.}\hspace{\namespacing}Dong~Tian\\
  InterDigital, New York, NY, USA\\
  {\tt\small jiahao.pang@interdigital.com, kevinb3@uci.edu, dong.tian@interdigital.com}
}

\begin{document}
\maketitle\copyrightnotice
\input{sec/0_abstract}

\section{Introduction}
\label{sec:intro}
\input{sec/1_intro}

\section{Related Work}
\label{sec:related}
\input{sec/2_related}

\section{Point-Voxel-Tree-based PCC Framework}
\label{sec:framework}
\input{sec/3_framework}

\section{Voxel Geometry Processing}
\label{sec:voxel}
\input{sec/4_voxel}

\section{Experimentation}
\label{sec:results}
\input{sec/5_results}

\vspace{-5pt}
\section{Conclusion}
\label{sec:conclusion}
\input{sec/6_conclusion}

{
\small
\bibliographystyle{ieeenat_fullname}

}

\end{document}

%% file: sec/0_abstract.tex
\begin{abstract}
The universality of the point cloud format enables many 3D applications, making the compression of point clouds a critical phase in practice.
Sampled as discrete 3D points, a point cloud approximates 2D surface(s) embedded in 3D with a finite bit-depth.
However, the point distribution of a practical point cloud changes drastically as its bit-depth increases, requiring different methodologies for effective consumption/analysis.
In this regard, a heterogeneous point cloud compression (PCC) framework is proposed.
We unify typical point cloud representations---point-based, voxel-based, and tree-based representations---and their associated backbones under a learning-based framework to compress an input point cloud at different bit-depth levels.
Having recognized the importance of voxel-domain processing, we augment the framework with a proposed context-aware upsampling for decoding and an enhanced voxel transformer for feature aggregation. 
Extensive experimentation demonstrates the state-of-the-art performance of our proposal on a wide range of point clouds.
\end{abstract}
\vspace{-10pt}

%% file: sec/1_intro.tex
As affordable depth-sensing devices develop, the point cloud format increasingly attracts interest in the ecosystem of 3D applications, such as AR/VR, robotics, autonomous driving, \etc\ 
A 3D point cloud describes the surface of an object or a scene by sampling a set of 3D points.
To cater to real-world usages in representation and analysis, a point cloud needs to store a huge number of points.
For instance, a 10-bit point cloud frame for AR/VR applications typically contains over a million of points. There is a practical demand for point cloud compression (PCC)~\cite{quach2022survey} to alleviate the transmission or storage of point cloud information.
In this work, we particularly focus on the lossy compression of point cloud geometry.

A point cloud to be compressed has finite precision indicated by its \emph{bit-depth}, the number of bits needed to describe the 3D coordinates of its points.
For example, a $10$-bit point cloud is a point set confined in a 3D box of size $1024\times 1024\times 1024$, where its $(x,y,z)$-coordinates are integers ranging between $0$ to $2^{10} - 1=1023$. In other words, the point cloud coordinates are represented by a $10$-bit binary number, where the first few bits correspond to the \emph{coarse} geometry/shape while the last few bits delineate the \emph{fine} details established on top of the coarse shape.
By removing the last few bits from the coordinates followed by removing the duplicate points, a point cloud is quantized to its coarser version.
For instance, removing the last $2$ bits from a $10$-bit point cloud results in an $8$-bit point cloud with coordinates ranging from $0$ to $255$.
Thus, the compression of a point cloud means to compress all bits of its points' coordinates, starting from its coarse representation (first few bits) to its finer details (last few bits).
With every additional bit that is compressed, the resolution of the whole 3D space increases by a factor of 2; meanwhile, the precision of the point cloud also increases by 2 times.

\begin{figure}[t]
  \centering\scriptsize
  \includegraphics[width=0.85\columnwidth]{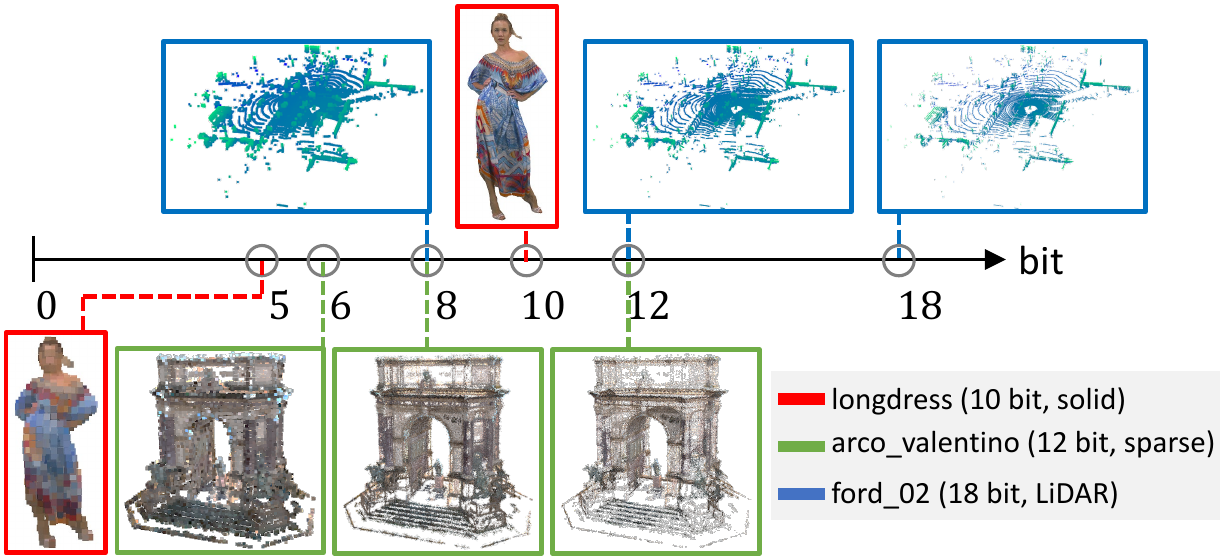}
  \caption{Several 3D point clouds at different bit-depth levels.}
  \label{fig:density}
  \vspace{-15pt}
\end{figure}

However, it is non-trivial to compress the bits of the point cloud coordinates.
As analyzed by \cite{pang2022grasp, wang2022sparse}, a practical point cloud has very different point distributions when it is inspected at different bit-depth levels, as demonstrated by the examples in Fig.~\ref{fig:density}.
When inspected at the first few bits, the points are \emph{densely distributed} in the space.
As bit-depth/granularity increases, distances between the 3D points gradually increase. At the last few bits representing the fine details, the 3D points become farther apart from one another.
In other words, the points are \emph{sparsely distributed}, as opposed to the distributions at the first few bits. Therefore, the varying point distributions request different methodologies/strategies to effectively process a point cloud at different bit-depths.

In this work, we unify the merits of different point cloud representations---the point-based, the voxel-based, and the tree-based representations---under one learning-based PCC framework that we call the PIVOT-Net (PoInt, VOxel and Tree).
It novelly considers the point distributions across bit-depth levels under the Rate-Distortion (RD) restriction of compression problems.
Similar to related works \cite{pang2022grasp, wang2021multiscale,quach2020improved}, for the first few bits where the neighboring points are highly correlated and compressible, a tree-based method is applied to compress the quantized version losslessly.
This lays a faithful foundation for compressing the remaining bits at a low cost.
For the last few bits where the 3D points are sparsely distributed and less correlated, we apply a point-based neural network to extract the local geometric features like in \cite{pang2022grasp}.

Different from \cite{pang2022grasp}, to compress the middle-range bits, where neighboring points are still considerably correlated, a 3D convolutional neural network (CNN) is applied for processing in the voxel domain.
Notably, PIVOT-Net augments the voxel-domain processing via a proposed context-aware upsampling for decoding/synthesis and via an enhanced voxel transformer for advanced feature aggregation.

The main contributions of the proposed PIVOT-Net include the following:
\begin{enumerate}[label=(\roman*)]
\item We propose the first learning-based PCC framework \emph{unifying} the point-based, voxel-based, and tree-based representations of point clouds to efficiently compress different bit-depth levels of a point cloud.
\item The voxel-domain processing is augmented by a proposed context-aware upsampling module and an enhanced voxel transformer module.
\item The proposed PIVOT-Net demonstrates the state-of-the-art compression performance on a wide spectrum of practical point clouds.
\end{enumerate}

%% file: sec/2_related.tex
Unlike an image whose pixels are naturally organized on a 2D grid, a point cloud contains \emph{unordered} 3D points in space. 
To consume the challenging point cloud data, several representations---point-based, voxel-based, and tree-based---are exploited.
We first review these representations, followed by illustrating how they facilitate the state-of-the-art PCC proposals.
To this end, related techniques for voxel-domain processing are discussed.

\textbf{Point cloud representations}: 
\emph{Voxel-based} representation organizes 3D points in the Euclidean domain, which enables them to be processed like images by 3D CNNs ~\cite{quach2022survey}.
It is achieved by uniformly quantizing the 3D coordinates to voxel grids with a given quantization step size $s$.
For example, a point $(x,y,z)$ is quantized to a voxel $(\lfloor x/s \rfloor, \lfloor y/s \rfloor, \lfloor z/s \rfloor)$ where $\lfloor\cdot\rfloor$ is the floor function.
Then for indication, a scalar ``$1$'' is assigned to the voxel to mark it as occupied while empty voxels are marked by ``$0$''~\cite{quach2022survey}.
During voxelization, a smaller $s$ creates many empty voxels that make the subsequent processing inefficient, whereas a larger $s$ compromises the representability of the geometry.
Thus, voxel-based representation is less suitable to delineate intricate details.

With a voxelized point cloud, the \emph{tree-based} representation views the occupied voxels as leaf nodes and organizes them under a hierarchical tree structure~\cite{huang2020octsqueeze}.
For instance, a voxelized point cloud with $10$-bit precision can be organized as an octree with depth level of $10$, where each level is represented by a bit.
Thus, encoding the whole octree means encoding the whole point cloud \emph{losslessly}; and encoding only the first few levels of the tree---the first few bits of the point cloud---is to encode the quantized version of the input losslessly~\cite{graziosi2020overview}.

\emph{Point-based} representation is a native point cloud representation, where a point cloud is simply a set of 3D points specified by their $(x, y, z)$-coordinates~\cite{quach2022survey,pang2022grasp}.
Unlike other representations, it does not require any pre-processing.
Thus, it accurately represents intricate details---the last few bits of a point cloud---without compromising known geometric information.
Though it is non-trivial to digest an unordered point set, recent progress in point-based deep neural networks has shed light on this problem.
The seminal work PointNet~\cite{qi2017pointnet} combines multi-layer perceptron (MLP) and pooling operators to extract permutation-invariant features. 
Inspired by the convolutional layer in Euclidean domain, subsequent works, such as PointCNN~\cite{li2018pointcnn} and PointConv~\cite{wu2019pointconv}, are proposed.
Point-based representation also applies to reconstruction. 
With a series of MLP layers, Latent~GAN~\cite{achlioptas2018learning} directly learns the 3D coordinates of a point cloud.
Other works, such as FoldingNet~\cite{yang2018foldingnet} and TearingNet~\cite{pang2021tearingnet}, embed a topology for reconstruction.

\begin{figure*}[t]
  \vspace{-10pt}
  \centering\scriptsize
  \captionsetup[subfigure]{width=270pt}
  \subfloat[Tree + voxel: PCGCv2\cite{wang2021multiscale}, SparsePCGC\cite{wang2022sparse}, and \cite{quach2020improved}]{\includegraphics[width=200pt]{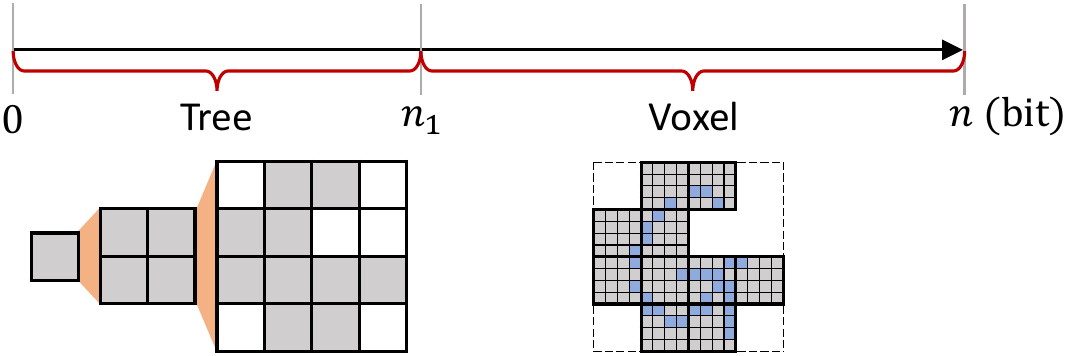}\label{fig:tv}}\hspace{30pt}\vspace{5pt}
  \subfloat[Tree + point]{\includegraphics[width=200pt]{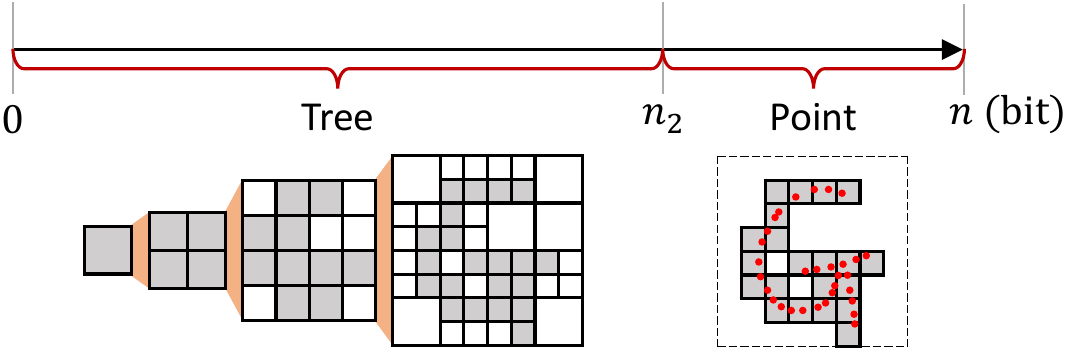}\label{fig:tp}}\\\vspace{-2pt}
  \subfloat[Tree + point,\,w/\,feat.\,down/upsampling:\,GRASP-Net~\cite{pang2022grasp}]{\includegraphics[width=200pt]{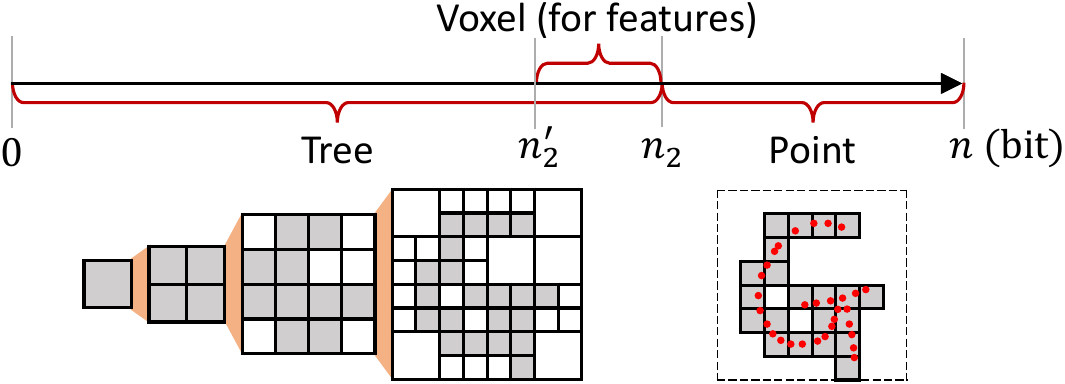}\label{fig:tvp}}\hspace{30pt}
  \subfloat[Tree + voxel + point,\,w/\,feat.\,down/upsampling:\,PIVOT-Net (Ours)]{\includegraphics[width=200pt]{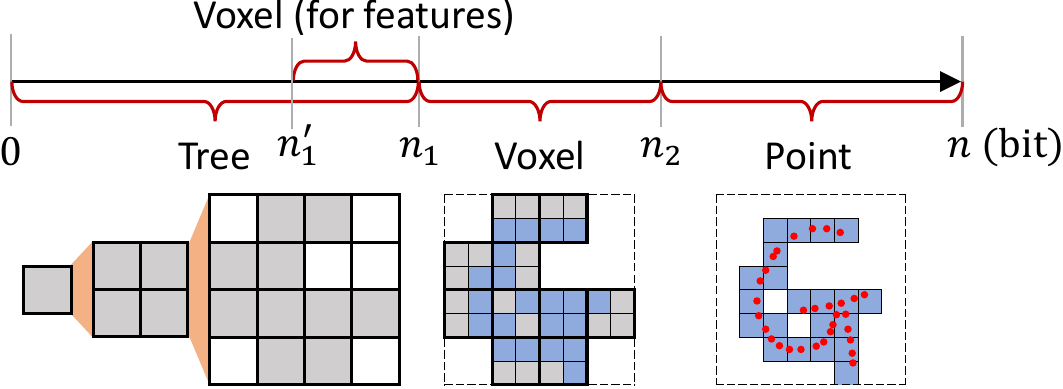}\label{fig:tvvp}}
  \caption{\small Comparisons of lossy PCC frameworks utilizing different point cloud representations.}
  \label{fig:comp}
  \vspace{-10pt}
\end{figure*}

\textbf{Point cloud compression}:
As the state-of-the-art, non-learning-based PCC method, MPEG G-PCC utilizes the octree representation for lossless point cloud coding~\cite{graziosi2020overview}.
It is achieved by a hand-crafted context model that predicts the voxel occupancy status, followed by an arithmetic coder for entropy coding.
Replacing its hand-crafted context model with a learnable one establishes a family of works called deep octree coding.
Representative works in this thread include OctSqueeze~\cite{huang2020octsqueeze} and VoxelContext-Net~ \cite{que2021voxelcontext}.

Based on the seminal work on end-to-end learning for image compression~\cite{balle2016end, balle2018variational}, another family of learning-based PCC methods is proposed to perform rate-distortion optimization for lossy point cloud geometry compression.
Different from deep octree coding that compresses occupancy status, this paradigm compresses geometric features generated by deep neural networks.
As first attempts of this thread, the works \cite{quach2019learning,quach2020improved,guarda2020adaptive,wang2021lossy} apply regular 3D CNN to compress voxelized point cloud.
Since most of the voxels are empty, regular 3D convolution is memory inefficient.
To counter this issue, PCGCv2~\cite{wang2021multiscale} and its follow-up effort SparsePCGC~\cite{wang2022sparse,wang2022sparse} apply 3D sparse CNN~\cite{choy20194d} for lossy compression based on an octree-coded coarse partitioning.

Compared to voxel representation and the associated CNN backbone, point-based representation is more effective for consuming local geometric details. 
However, there is limited progress in using point-based representation for learning-based PCC.
Many existing works, such as \cite{yan2019deep,huang20193d,you2021patch,wiesmann2021deep,al2022variable}, fail to justify their performance on real-world data, \eg, the MPEG group's recommended test point clouds~\cite{ctcgpcc}.
The recently proposed GRASP-Net~\cite{pang2022grasp} has successfully used point-based learning and octree coding for promising compression performance.
However, GRASP-Net is \emph{sub-optimal}.
It only digests the very last few bits (\eg, the last 2 bits for a 12-bit point cloud) with a point-based neural network, while the rest (\eg, the first 10 bits for a 12-bit point cloud) are coded losslessly with an octree coder that can be very inefficient.

\textbf{Voxel-domain processing}:
Unlike GRASP-Net, in this work, we relieve the burden of the octree coder by introducing additional voxel-based processing to consume the middle-range bits.
The aforementioned works PCGCv2~\cite{wang2021multiscale} and SparsePCGC~\cite{wang2022sparse} utilize sparse CNNs to achieve upsampling, followed by binary classification and pruning in the voxel domain.
Following a similar methodology, while being inspired by the context modeling in octree coding~\cite{huang2020octsqueeze,que2021voxelcontext}, we propose an adaptive voxel upsampling process for decoding that is \emph{context aware}.

Feature aggregation also plays a critical role in the processing and analysis of voxel geometry.
In \cite{quach2020improved}, Quach~\etal\ applies the classic ResNet~\cite{he2016deep} for analysis and synthesis in the voxel domain.
Wang~\etal\ \cite{wang2021multiscale,wang2022sparse} then extend the Inception ResNet (IRN) proposed for 2D images \cite{szegedy2017inception} to 3D.
The recently proposed \emph{vision transformer} module~\cite{dosovitskiy2020image} for 2D image analysis adaptively counts the contributions from nearby pixels according to a self-attention model, which enables a large receptive field with affordable computational cost.
It is later adapted for point cloud processing, leading to point transformer~\cite{zhao2021point} and voxel transformer~\cite{mao2021voxel}. 
In the PIVOT-Net, we enhance the voxel transformer for advanced geometric feature aggregation.

%% file: sec/3_framework.tex
First we analyze typical lossy PCC frameworks to understand how they utilize different point cloud representations.
Then we introduce the proposed PIVOT-Net that combines point-based, voxel-based, and tree-based representations.

\subsection{Heterogeneous PCC Frameworks}
\textbf{Overview}: Fig.\,\ref{fig:comp} compares several lossy PCC frameworks to encode an $n$-bit point cloud.
For each framework, different point cloud representations are applied to compress different intervals along the bit axis.
First and foremost, in all frameworks, a \emph{tree-based} (\eg, octree-based) coder is first applied to code the first few bits losslessly as a bit-stream, which results in a coarse, quantized representation of the input point cloud.
For instance, see the occupied leaf voxels (in gray) of the tree at the bottom-left of Fig.\,\ref{fig:tv} where the point cloud is partitioned as occupied blocks at the bit-depth level of $n_1$.
And this \emph{partitioning} information is losslessly recorded by compressing the octree. 
Thus, the associated bit-stream of the octree is called the \emph{partitioning} bit-stream.
After partitioning the point cloud into blocks, how do we encode the point coordinates within each block? In other words, how do we encode the remaining bits?

\textbf{Tree+voxel}: Under the general tree+voxel design in Fig.~\ref{fig:tv}, several works, including PCGCv2~\cite{wang2021multiscale}, SparsePCGCv1~\cite{wang2022sparse}, and \cite{quach2020improved}, view the points within each block as voxels, allowing them to apply a CNN to extract a geometric feature for each block.
Unlike traditional compression paradigm (\eg, HEVC~\cite{sullivan2012overview}) where information rarely flows across block boundaries, utilizing deep neural networks naturally aggregates information across blocks.
Depending on the receptive field of the networks, the generated feature $\mathbf{f}_A$ associated with the block $A$ may not only abstract the geometry of $A$ but also contain information from other blocks surrounding $A$. 
Hence, CNNs can uncover the underlying correlation between nearby voxels to generate compact, informative features that can facilitate better coding performance.
The extracted features are organized as a 3D feature map of size $2^{n_1}\times2^{n_1}\times2^{n_1}$, where $n_1$ is the number of bits already processed in tree-based representation. 
The nonempty entries of the feature map only correspond to the occupied blocks (in gray)  as they are capable of generating features.
The feature map is then entropy coded as another bit-stream that we call the \emph{feature} bit-stream to represent the geometry of the last $n-n_1$ bits.

\textbf{Tree+point}: However, the tree+voxel design fails on sparse point cloud~\cite{pang2022grasp} such as LiDAR sweeps because the inherent lack of neighbors undermines the performance of CNNs.
To address this problem, the points of each block are represented as raw 3D points in $(x,y,z)$-coordinates so that they are fed into \emph{point-based} neural networks to extract blockwise features.
This process leads to the tree+point design in Fig.~\ref{fig:tp}.
This design is motivated by the fact that point-based neural networks are flexible in representing intricate details without heavily relying on neighboring points, as mentioned in Section~\ref{sec:related}.
Extracting descriptive features with point-based networks should only be applied to the very last few bits where the point cloud gets very sparse.
Thus, compared with the tree+voxel design, the tree+point design requires more bits of the point cloud to be coded in tree-based representation. In other words, $n_2$ in Fig.~\ref{fig:tp} is larger than $n_1$ in Fig.~\ref{fig:tv}, leaving a smaller bit interval to be coded in point-based representation.

\textbf{GRASP-Net}~\cite{pang2022grasp}: The tree+point design is not optimal for PCC. 
Compared with the tree+voxel design, the tree+point design requires a larger bit-stream because of two reasons: (i)~the feature map generated from the last few bits is finer, and (ii)~the tree-based representation is deeper.
To resolve (i), GRASP-Net additionally downsamples (or upsamples when decoding) the feature map in the voxel domain by a CNN, as shown in Fig.~\ref{fig:tvp}.
Via downsampling with a CNN, the resolution of the feature map is reduced from $2^{n_2}\times 2^{n_2}\times 2^{n_2}$ to $2^{n'_2}\times 2^{n'_2}\times 2^{n'_2}$ (when encoding) where $n'_2 < n_2$.
The downsampled feature map is further entropy coded.
In this design, the point cloud geometry is not consumed by the CNN.
Differently, the known geometry at bit-depth level $n_2$ serves as a support to down-/up- sample the features attached to the point cloud blocks.

\textbf{Our proposal}: Note that GRASP-Net still encodes the finer-grained block partitioning of the first $n_2$ bits.
Motivated by the CNN in the tree+voxel design (Fig.~\ref{fig:tv}) for digesting geometry, we propose PIVOT-Net as the first framework for the \emph{unified} tree+voxel+point design in Fig.~\ref{fig:tvvp}.
Instead of handling bit interval $[n_1, n_2]$ with an octree coder in GRASP-Net, the proposed PIVOT-Net represents these middle-range bits with voxels and consumes them with a CNN, thereby reducing the burden of the octree coder.

Based on a novel processing of the fine, middle and coarse levels of point clouds with the point-, voxel-, and tree-based modules, our PIVOT-Net considers the point distributions across bit-depth levels under the Rate-Distortion (RD) restriction in compression.
Thus, each main component of PIVOT-Net \emph{can be configured} to handle a specific range of point distribution to achieve superior PCC.
\begin{figure}
  \centering
  \includegraphics[width=0.9\columnwidth]{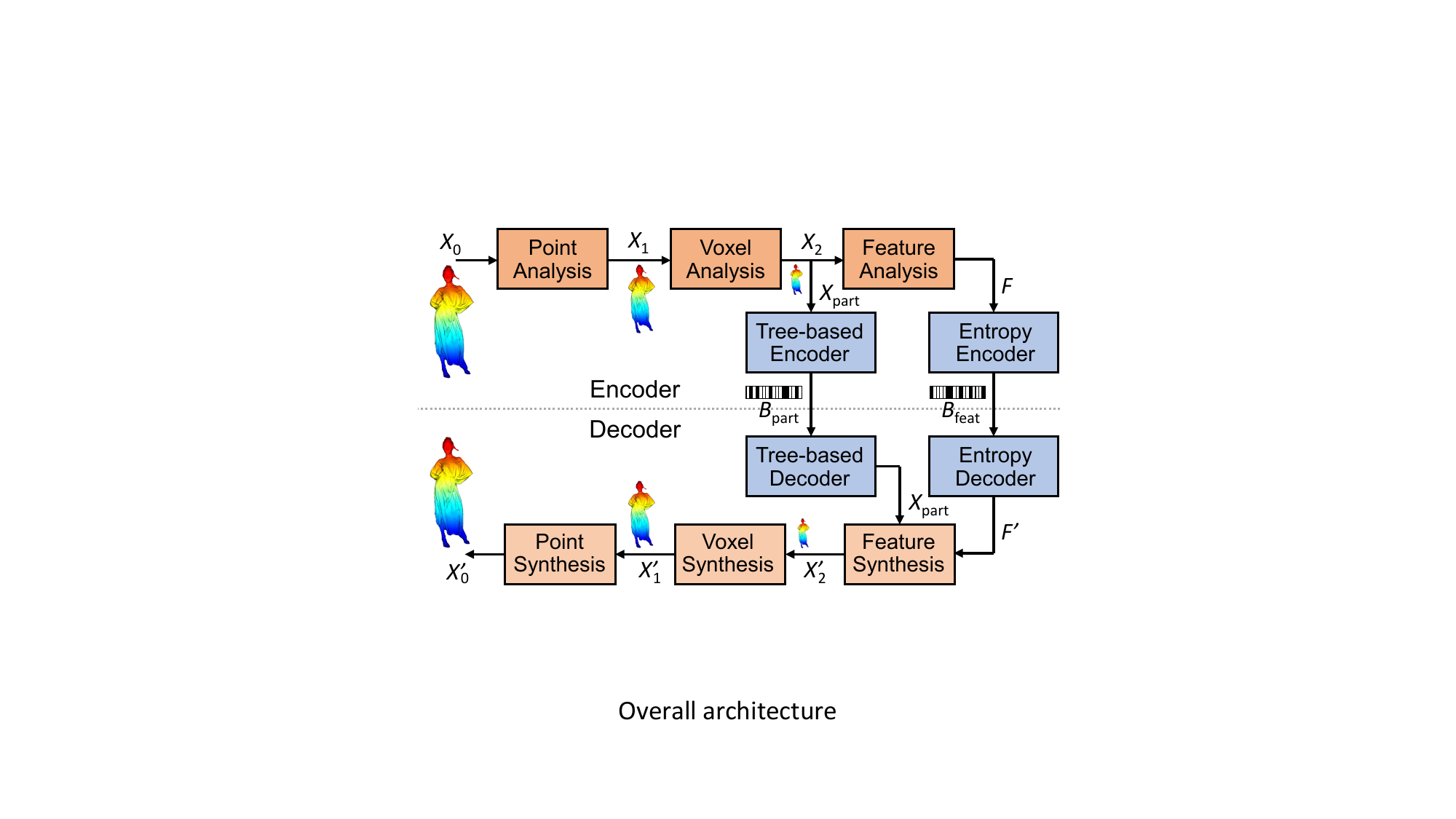}
  \caption{Architecture of our PIVOT-Net. The orange blocks contain learnable neural network layers, where point analysis/synthesis are point-based neural networks while voxel and feature analysis/synthesis are sparse CNNs.}
  \label{fig:pivot}
  \vspace{-10pt}
\end{figure}

\subsection{The PIVOT-Net Architecture}\label{ssec:pivot}
The block diagram of the PIVOT-Net is shown in Fig.~\ref{fig:pivot}, where the neural network modules are colored in orange.

\textbf{Encoder}:
An input point cloud $X_0$ is fed into the \emph{Point Analysis Network}, a point-based neural network akin to PointNet++~\cite{qi2017pointnet++}.
In contrast to PointNet++ relying on farthest point sampling to obtain a coarse point cloud, the Point Analysis Network first generates the coarse representation $X_1$ by uniform quantization with a constant step size $s_1>1$, where $n-n_2 = \log_2 s_1$.
Then with the set abstraction layer from PointNet++, it extracts a geometric feature for each point in $X_1$ with the nearest neighbor search followed by a shared PointNet.
Thus, the last $\log_{2}{s_1}$ bits representing the fine geometric details are consumed with this point-based network (the bit interval $[n_2, n]$ of Fig.~\ref{fig:tvvp}).

Next, in the 3D tensor format, the coarse point cloud $X_1$ (appended with the geometric features generated from the Point Analysis Network) is fed into the \emph{Voxel Analysis Network} that uses a sparse CNN for downsampling and feature aggregation similarly done in PCGCv2~\cite{wang2021multiscale}.
By downsampling $X_1$ by $s_2$ times to obtain the output $X_2$ and its features, the Voxel Analysis Network abstracts from $X_1$ the last $\log_{2}{s_2}$ bits, which corresponds to the bit interval $[n_1, n_2]$ in Fig.~\ref{fig:tvvp}, \ie, $n_2-n_1=\log_{2}{s_2}$.

The voxel geometry, or occupancy status, of $X_2$ is denoted as $X_\textrm{part}$, which contains the block partitioning information.
$X_\textrm{part}$ is a coarse version of the input $X_0$ with an overall quantization step size $s=s_1s_2$.
It is losslessly coded with the tree-based encoder as a partitioning bit-stream $B_\textrm{part}$, corresponding to the bit interval $[0, n_1]$ in Fig.~\ref{fig:tvvp}.
Moreover, the feature map of $X_2$ shrinks by $s_3$ times using the \emph{Feature Analysis Network}, another sparse CNN for downsampling.
Thus, the Feature Analysis Network is applied to the bit interval $[n'_1,n_1]$ in Fig.~\ref{fig:tvvp}, \ie, $n_1-n'_1=\log_{2}{s_3}$.
The downsampled feature map $F$ is finally entropy encoded as the feature bit-stream $B_\textrm{feat}$.

\textbf{Decoder}:
The partitioning information $X_\textrm{part}$ is decoded from the partitioning bit-stream $B_\textrm{part}$.
Meanwhile, the feature bit-stream $B_\textrm{feat}$ is decoded for the (decoded) downsampled feature map $F'$.
Next, both $X_\textrm{part}$ and $F'$ are fed to the \emph{Feature Synthesis Network}, a sparse CNN similar to the one in GRASP-Net~\cite{pang2022grasp}.
It upsamples the input feature map $F'$ while refining it to match the occupancy status of $X_\textrm{part}$, leading to an output tensor $X'_2$ with the upsampled feature map attached to it.
Note that $X'_2$ and $X_2$ have different features, but they share the same geometry as $X_\textrm{part}$. 

$X'_2$ is then fed into the \emph{Voxel Synthesis Network}.
Like the Feature Synthesis Network, the Voxel Synthesis Network is also a sparse CNN for upsampling.
However, while the Feature Synthesis Network intends to upsample the features $F'$ to a known geometry $X_\textrm{part}$, the Voxel Synthesis Network aims at upsampling the \emph{geometry} of $X'_2$ based on its features.
Outputted from the Voxel Synthesis Network, $X'_1$ has the same resolution as $X_1$, \ie, $2^{n_2}\times2^{n_2}\times2^{n_2}$.

In the end, consisting of a series of multi-layer perceptron (MLP) layers, the \emph{Point Synthesis Network} \cite{pang2022grasp}  recovers the raw 3D points for the final reconstruction. For each geometric feature in $X'_1$, 
it generates a 3D point set that delineates the local geometric details in the neighborhood of the point cloud block $A$.
The decoded point cloud $X'_0$ is the aggregation of all the 3D point sets generated by the Point Synthesis Network.
Please refer to the supplementary material for additional details on architecture designs.

%% file: sec/4_voxel.tex
Our proposed PIVOT-Net stands out among the other designs in Fig.~\ref{fig:comp} because it utilizes voxel-domain processing for consuming the middle-range geometric bits.
This section illustrates our efforts in capitalizing voxel-domain processing. 
We incorporate two major components for voxel-domain processing: (i)~a context-aware upsampling process for adaptive voxel synthesis and (ii)~an Enhanced Voxel Transformer inspired by \cite{mao2021voxel} for feature aggregation.
\begin{figure}
  \centering
  \includegraphics[width=1.0\columnwidth]{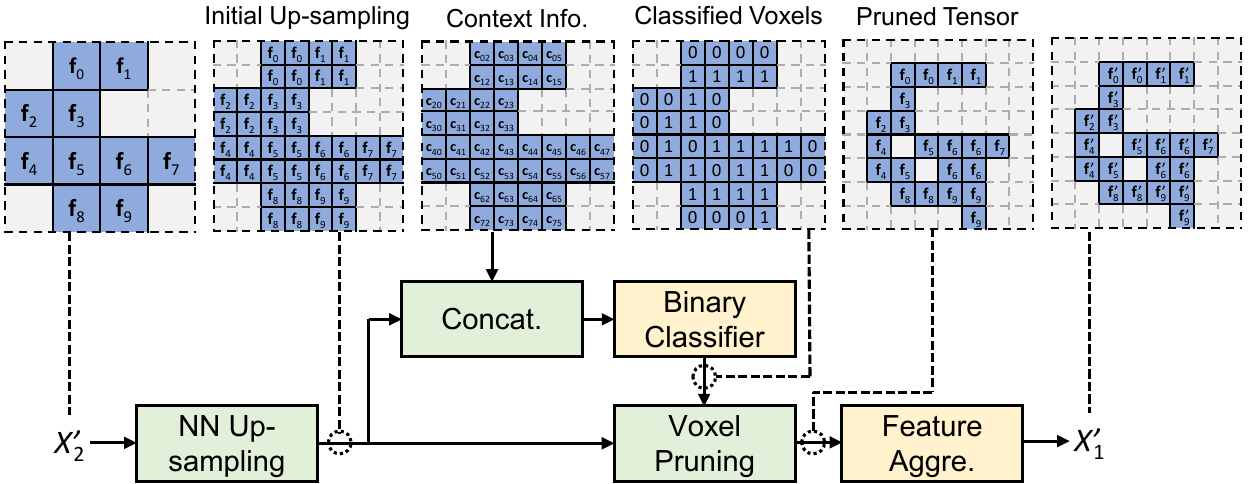}
  \vspace{-15pt}
  \caption{Context-aware upsampling for adaptive voxel synthesis. Learnable modules are colored in yellow.}
  \label{fig:voxup}
  \vspace{-15pt}
\end{figure}
\subsection{Adaptive Voxel Synthesis}
Downsampling and feature aggregation are straightforward to perform with sparse CNN, such as using convolutional layers with stride $2$ as done in \cite{quach2020improved} and \cite{wang2021multiscale}.
For simplicity, our work adopts the ``down-scale'' module from PCGCv2~\cite{wang2021multiscale} for the Voxel Analysis Network.

On the other hand, upsampling in the voxel domain is non-trivial.
The na\"ive nearest-neighbor (NN) upsampling always splits a parent voxel into $8$ child voxels, which leads to not only inaccurate upsampled geometry but also a waste of computation because many irrelevant occupied voxels are created.
To address these issues, PCGCv2~\cite{wang2021multiscale} and SparsePCGC~\cite{wang2022sparse} perform binary classification on the newly created child voxels and remove some of them that are less likely to be occupied in the ground truth.

Inspired by the \emph{context modeling} in octree coding, we additionally utilize the known knowledge about the child voxels to determine which child voxels should be pruned and which to be kept.
The block diagram of the proposed Voxel Synthesis Network is shown in Fig~\ref{fig:voxup}.
For simplicity, we assume the voxel analysis/synthesis network digests only $1$ bit.
Thus, the voxel synthesis network performs upsampling by a factor of $2$, though it can be cascaded a few times if more middle-range bits are coded.

To upsample the point cloud $X_2'$ and its associated feature map, they are first fed to an NN upsampling module using transposed convolution \cite{gwak2020gsdn} to obtain an initial upsampled tensor that is two times larger along each dimension.
After that, they are concatenated with the context information, followed by being fed to a binary classification module~\cite{wang2021multiscale} to determine which voxels should be removed.
In this work, the context information of a child voxel includes its $(x,y,z)$-coordinate and the current bit-depth level.
Next, the geometry of the initial upsampled point cloud is refined using voxel pruning according to the classification output.
In the end, a feature aggregation module with sparse convolutions is appended to improve the features based on the refined geometry, resulting in the output point cloud $X_1'$.

\subsection{Enhanced Voxel Transformer}
Compared to point-based neural network that usually has a large receptive field for capturing neighboring relationships, voxel-based processing with convolutional layers appears to be more ``rigid'' as it is unable to discover long-range dependency, especially for sparse point clouds.
However, CNN-based feature aggregation, \eg, with 3D Inception ResNet (IRN)~\cite{wang2021multiscale,pang2022grasp}, in the voxel-domain is still critical for generating descriptive geometric features.
To overcome this inherent issue of CNN-based feature aggregation, we propose the \emph{Enhanced Voxel Transformer} inspired by \cite{mao2021voxel} as a better alternative for feature aggregation.

\begin{figure}
  \centering
  \includegraphics[width=0.75\columnwidth]{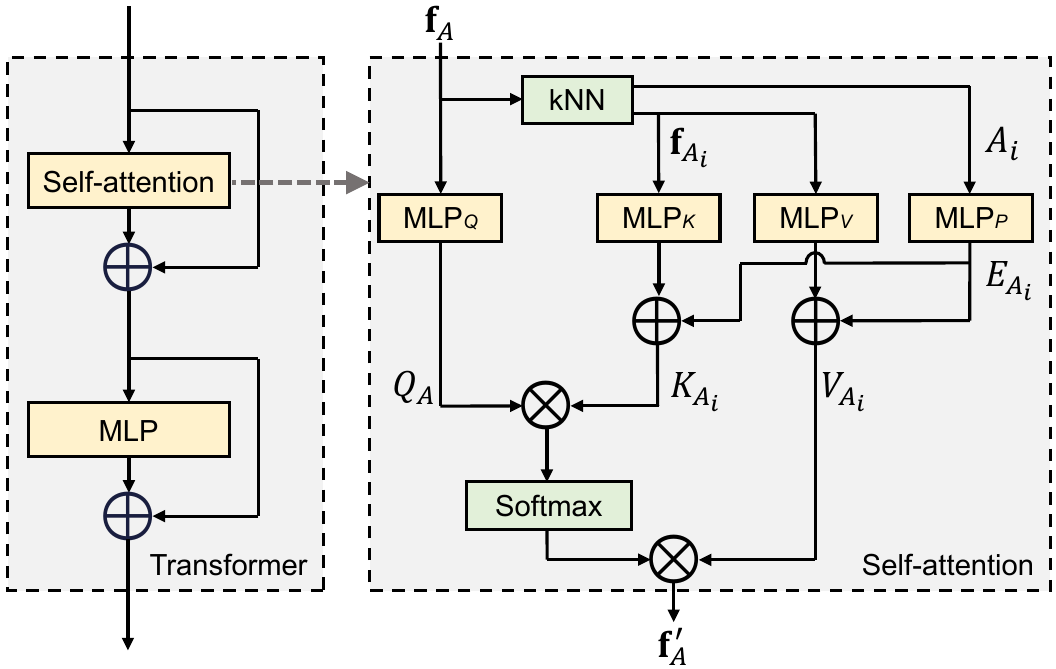}
  \vspace{-6pt}
  \caption{Enhanced Voxel Transformer (left) and its self-attention module (right). Learnable modules are colored in yellow.}
  \label{fig:trans}
  \vspace{-13pt}
\end{figure}

An Enhanced Voxel Transformer block is shown in the left of Fig.~\ref{fig:trans}, which consists of a self-attention block and an MLP block.
The self-attention block is detailed in the right of Fig.~\ref{fig:trans}.
Given a current feature vector $\mathbf{f}_A$ associated with a voxel $A$ and its neighboring $k$ features $\mathbf{f}_{A_i}$'s associated with the $k$ nearest voxels $A_i$'s, the self-attention block updates $\mathbf{f}_A$ based on the neighboring features $\mathbf{f}_{A_i}$'s using learnable MLP blocks $\textrm{MLP}_Q$, $\textrm{MLP}_K$, $\textrm{MLP}_V$ and $\textrm{MLP}_P$.

Firstly, the query embedding of $A$ is obtained via $Q_A=\textrm{MLP}_Q(\mathbf{f}_A)$.
The key embedding $K_{A_i}$ and the value embedding $V_{A_i}$ of all the nearest neighbors of $A$ are computed as
\begin{equation}
K_{A_i}\hspace{-1pt}=\hspace{-1pt}\textrm{MLP}_K(\mathbf{f}_{A_i})\hspace{-1pt}+\hspace{-1pt}E_{A_i}, V_{A_i}\hspace{-1pt}=\hspace{-1pt}\textrm{MLP}_V(\mathbf{f}_{A_i})\hspace{-1pt}+\hspace{-1pt}E_{A_i},
\end{equation}
for $0\le i\le k-1$. The term $E_{A_i}$ is the positional encoding between the voxels $A$ and $A_i$ calculated by
\begin{equation}
E_{A_i}=\textrm{MLP}_P(P_A-P_{A_i}),
\end{equation}
where $P_A$ and $P_{A_i}$ are the 3D coordinates of the voxels $A$ and $A_i$, respectively.
Then the updated feature of $A$ is 
\begin{equation}
\mathbf{f}'_A=\sum\nolimits_{i=0}^{k-1} \sigma\left(Q_A^\textrm{T}K_{A_i}\Big/c\sqrt{d}\right)\cdot V_{A_i},
\end{equation}
where $\sigma(\cdot)$ is the softmax function for computing the attention map, $d$ is the length of $\mathbf{f}'_A$, and $c$ is a constant.

In contrast to \cite{mao2021voxel}, we replace the linear projection layers with MLP layers for more flexibility when computing the attention map and the value embedding.
In this work, we use the Enhanced Voxel Transfomer blocks for feature aggregation in the Feature Synthesis Network (Fig.~\ref{fig:pivot}) to have a good trade-off between complexity and performance, where we cascade $3$ blocks with \emph{shared} weights.
Empirically, we find that our weight-shared transformers lead to a smaller model with better results.

%% file: sec/5_results.tex
\vspace{-2pt}
\subsection{Experimental Setup}
\vspace{-5pt}
\textbf{Datasets}:
Compression of the PIVOT-Net is verified on a comprehensive set of selected point clouds suggested by the MPEG group for learning-based PCC~\cite{mpeg2023dataset}.
With the taxonomy of \cite{mpeg2023dataset}, selected point clouds are categorized into four types---(i)~$4$ \emph{solid} surface point clouds ($10$--$11$ bits), (ii)~$3$ \emph{dense} surface point clouds ($12$ bits), (iii)~$5$ \emph{sparse} surface point clouds ($12$--$13$ bits), and (iv)~``ford\_02'' \& ``ford\_03'' \emph{LiDAR} sequences with $3000$ point cloud frames collected by a spinning LiDAR ($18$ bits).
The surface point clouds have a number of points ranging from $272K$ to $4.8M$; the LiDAR point clouds have about $80K$ points.
Example point clouds in \cite{mpeg2023dataset} are visualized in Fig.~\ref{fig:density}.

To compress surface point clouds, PIVOT-Net is trained with the ModelNet40~\cite{wu20153d} dataset which contains 12k CAD models from $40$ categories of objects.
As recommended by \cite{mpeg2023dataset}, for the case of LiDAR point clouds, we train the PIVOT-Net with the LiDAR sequence ``ford\_01'' containing $1500$ point cloud frames.

\textbf{Implementation details}:
PIVOT-Net is implemented based on \emph{PccAI}~\cite{pccai}---a testbed for learning-based PCC.
It is then  end-to-end trained with the rate-distortion (R-D) loss $L = L_\mathrm{D} + \lambda L_\mathrm{R}$.
Here $L_\mathrm{D}$ measures the geometric distortion and $L_\mathrm{R}$ is the estimated bitrate of $B_\textrm{feat}$ while $\lambda$ controls their trade-off.
The distortion is computed as $L_\mathrm{D}=\alpha L_\mathrm{CD}+\beta L_\mathrm{BCE}$, where $L_\mathrm{CD}$ is the augmented Chamfer distance between the ground-truth and the decoded point clouds \cite{yang2018foldingnet, chen20203d, pang2022grasp} and $L_\mathrm{BCE}$ is the average binary cross entropy loss between the classification output (before thresholding) and the voxelized ground-truth point cloud.

\def\rd_hght{105pt}
\begin{figure*}[htbp]
  \centering \scriptsize
  \includegraphics[width=1.4\columnwidth]{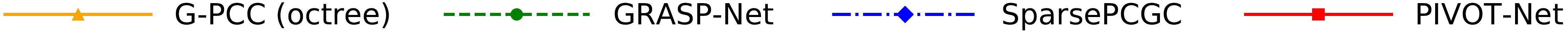}\vspace{4pt}\\
  \subfloat{\includegraphics[height=\rd_hght]{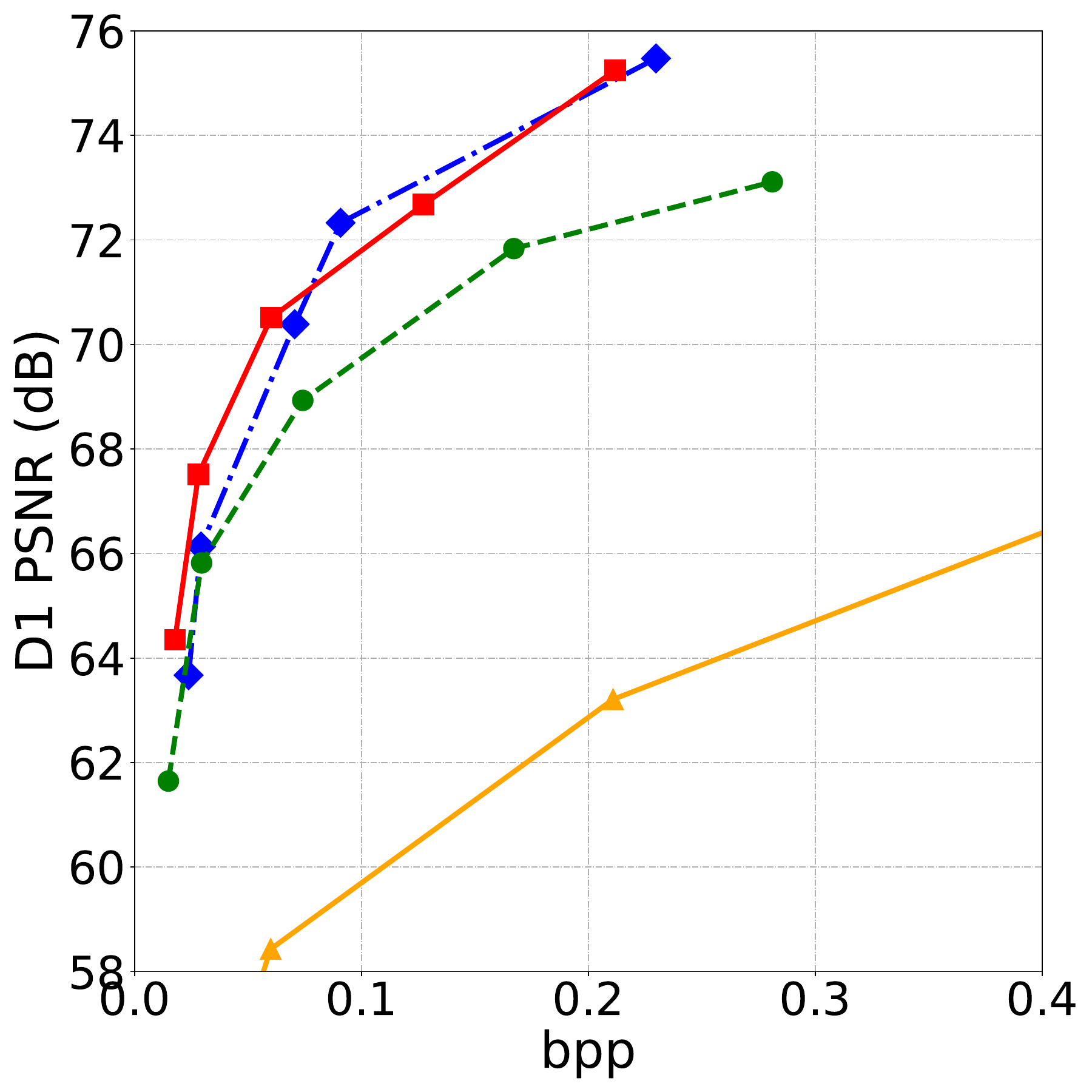}}\hspace{3pt}
  \subfloat{\includegraphics[height=\rd_hght]{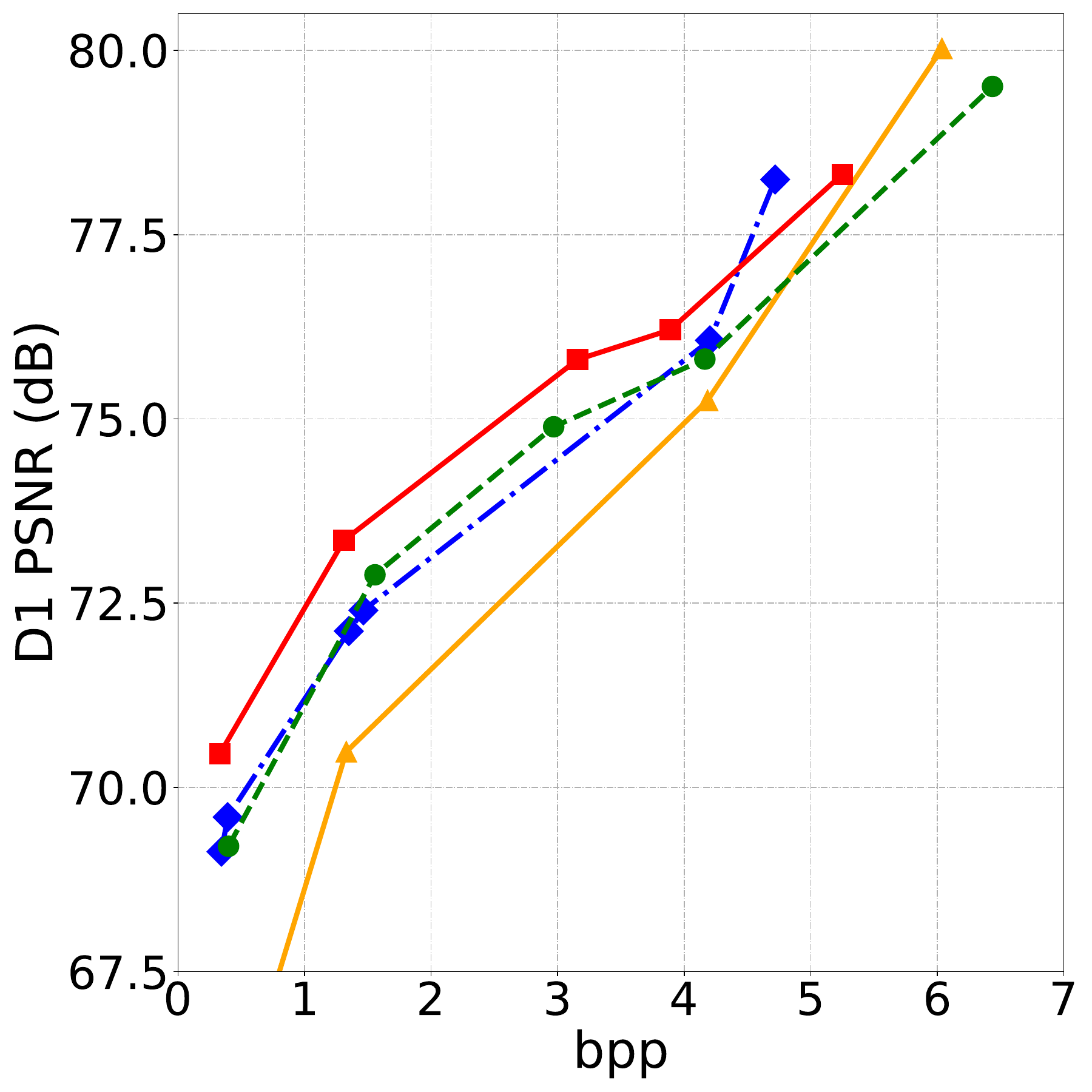}}\hspace{3pt}
  \subfloat{\includegraphics[height=\rd_hght]{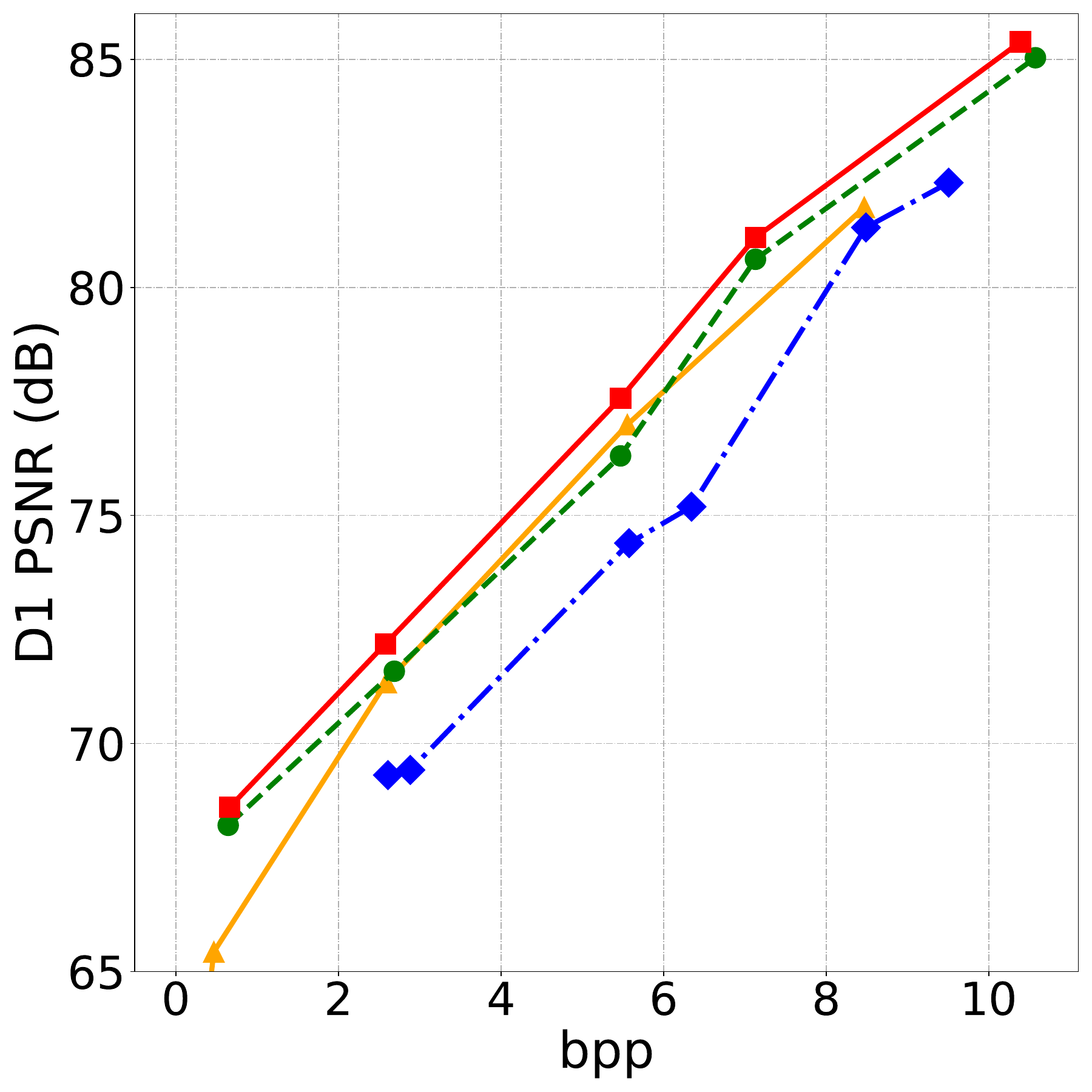}}\hspace{3pt}
  \subfloat{\includegraphics[height=\rd_hght]{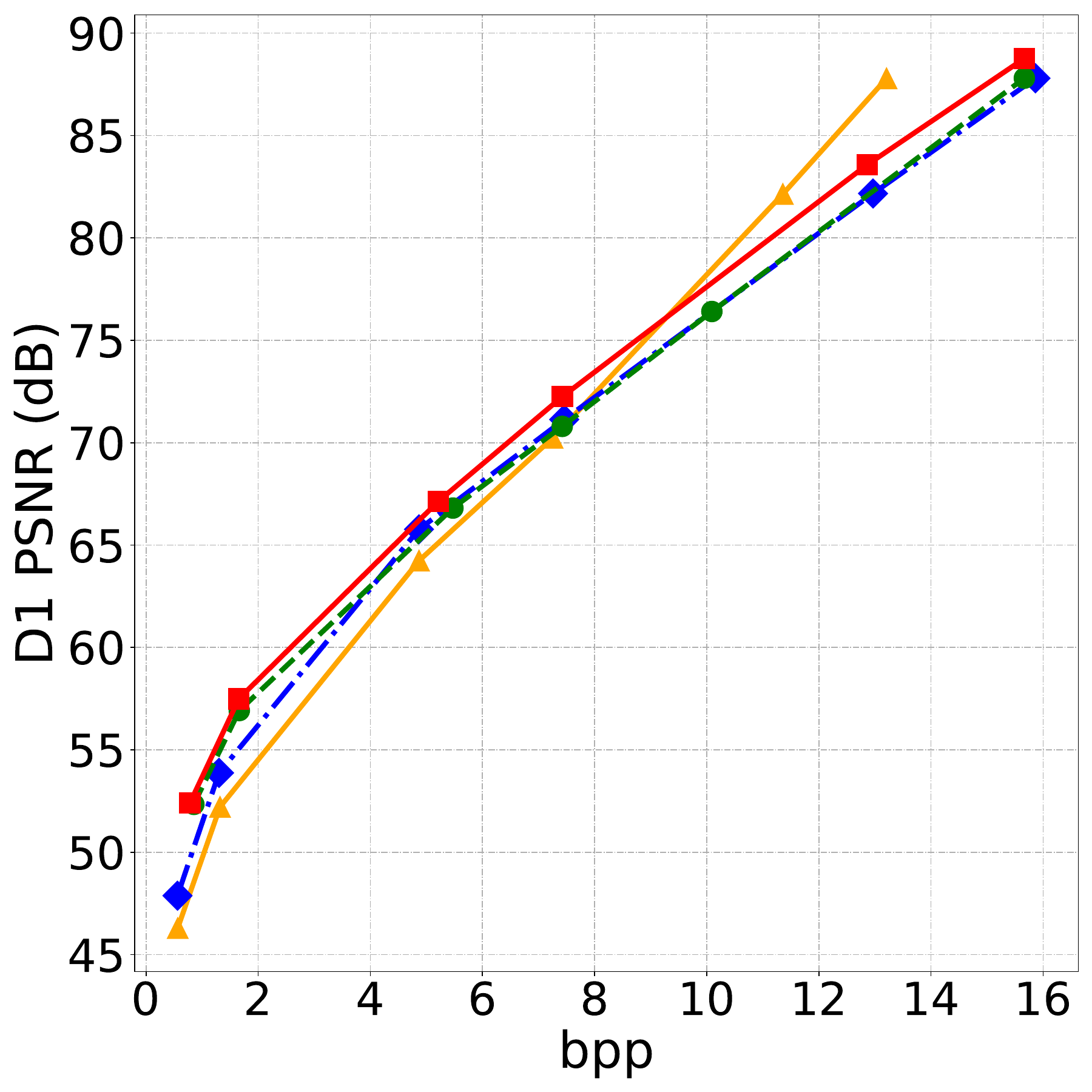}}\\\vspace{-2pt}
  \setcounter{subfigure}{0}
  \subfloat[``queen\_0200'' (solid)]{\includegraphics[height=\rd_hght]{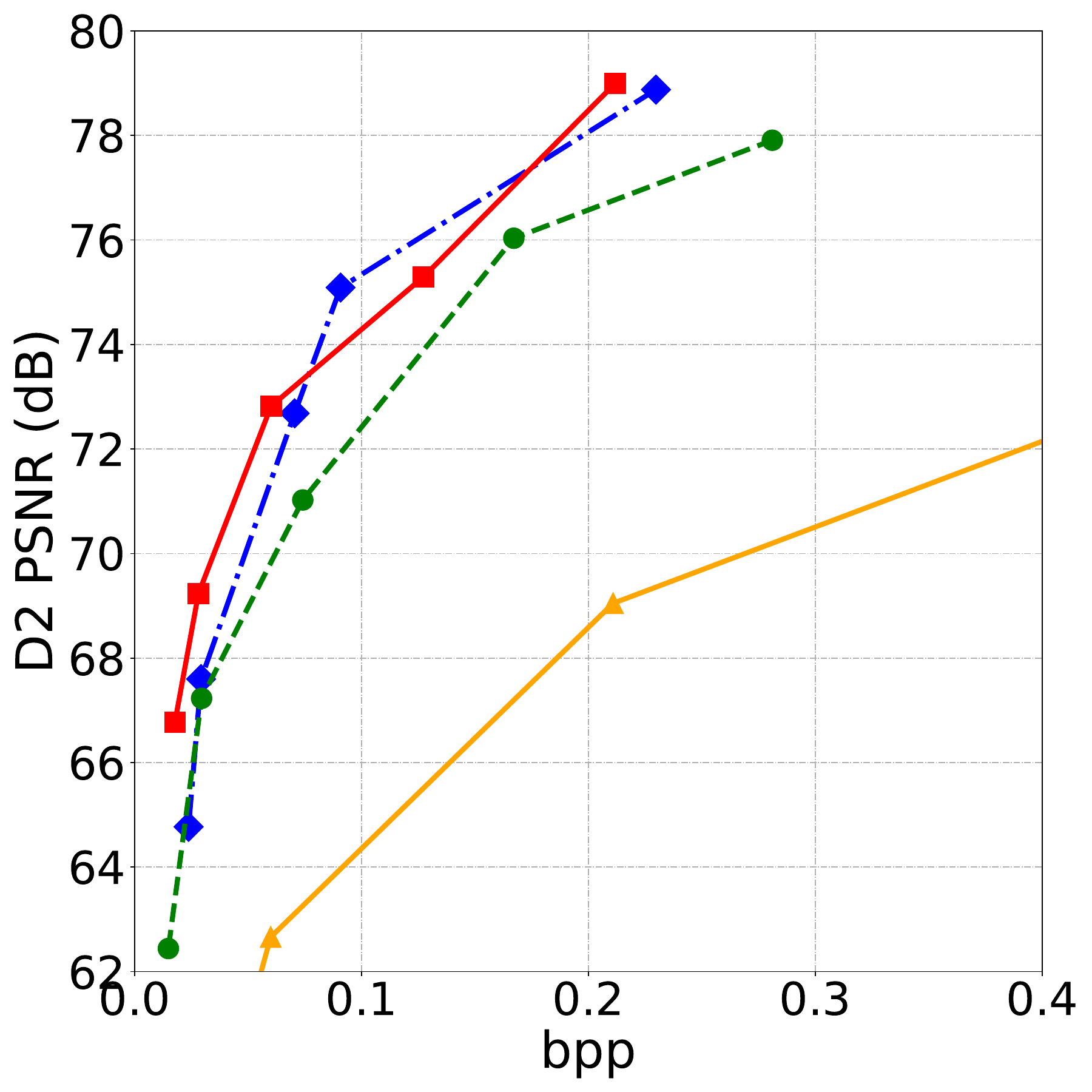}\label{fig:rd_solid}}\hspace{3pt}
  \subfloat[``facade\_09\_vox12'' (dense)]{\includegraphics[height=\rd_hght]{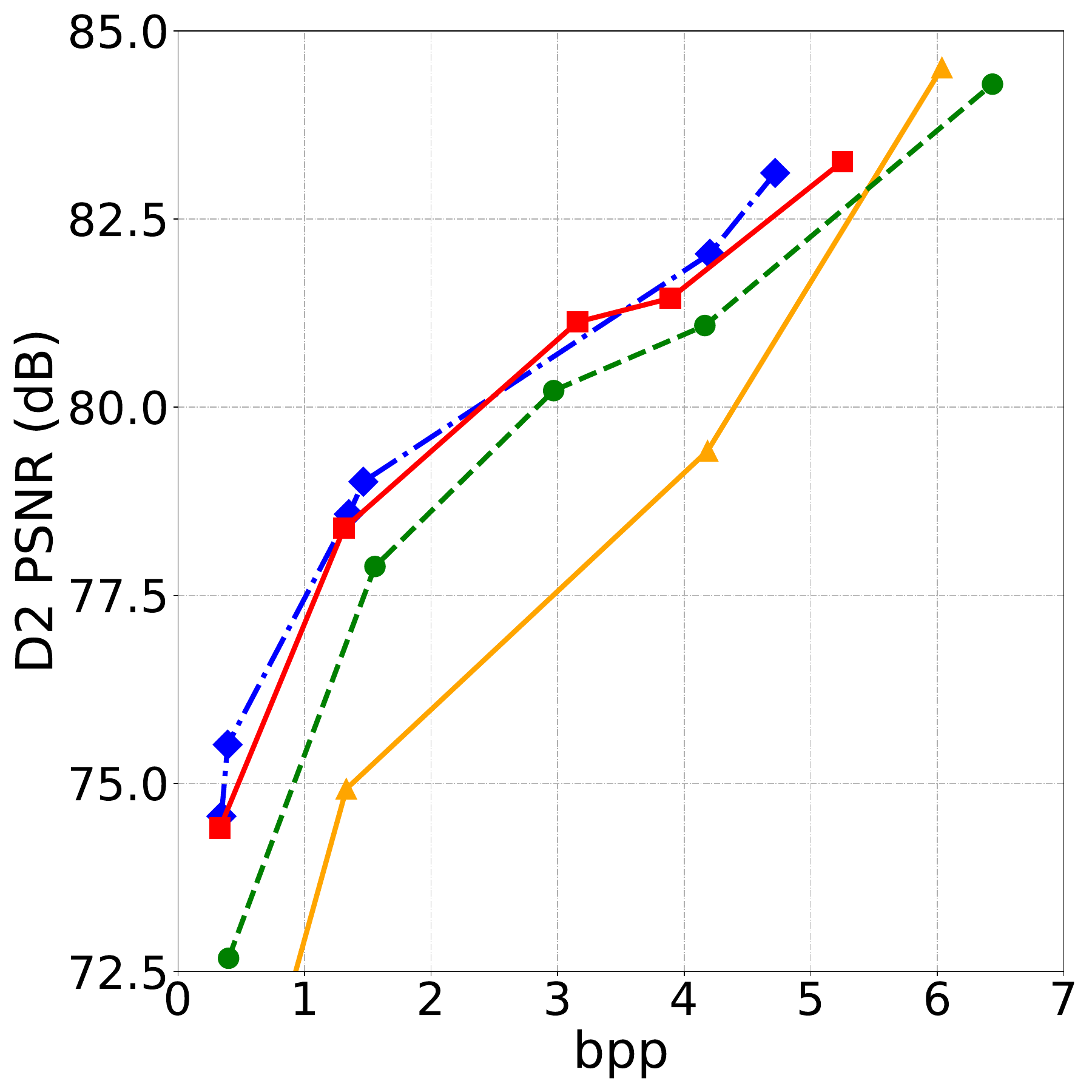}\label{fig:rd_dense}}\hspace{3pt}
  \subfloat[``ulb\_unicorn\_vox13'' (sparse)]{\includegraphics[height=\rd_hght]{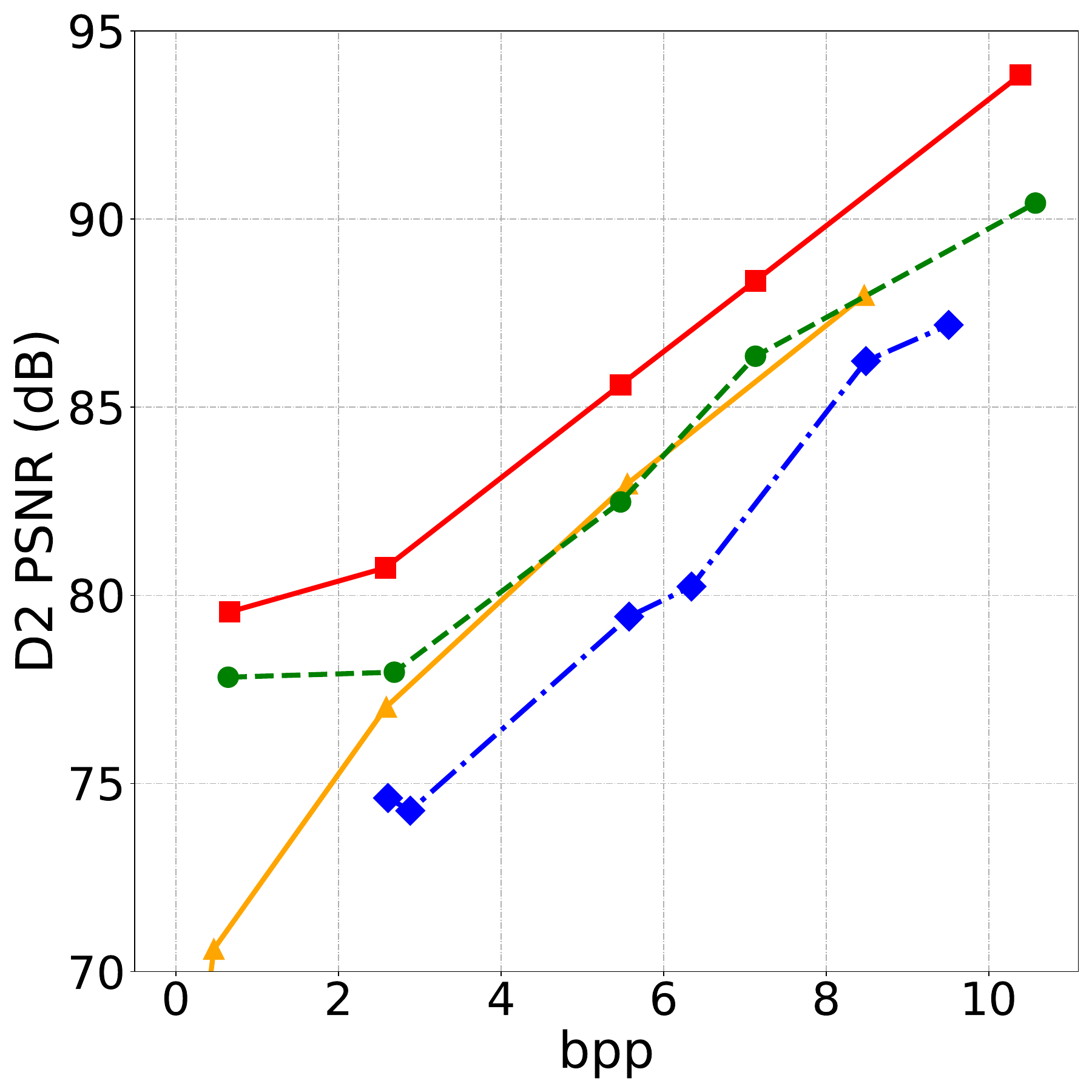}\label{fig:rd_sparse}}\hspace{3pt}
  \subfloat[``ford\_02\_q1mm'' (LiDAR)]{\includegraphics[height=\rd_hght]{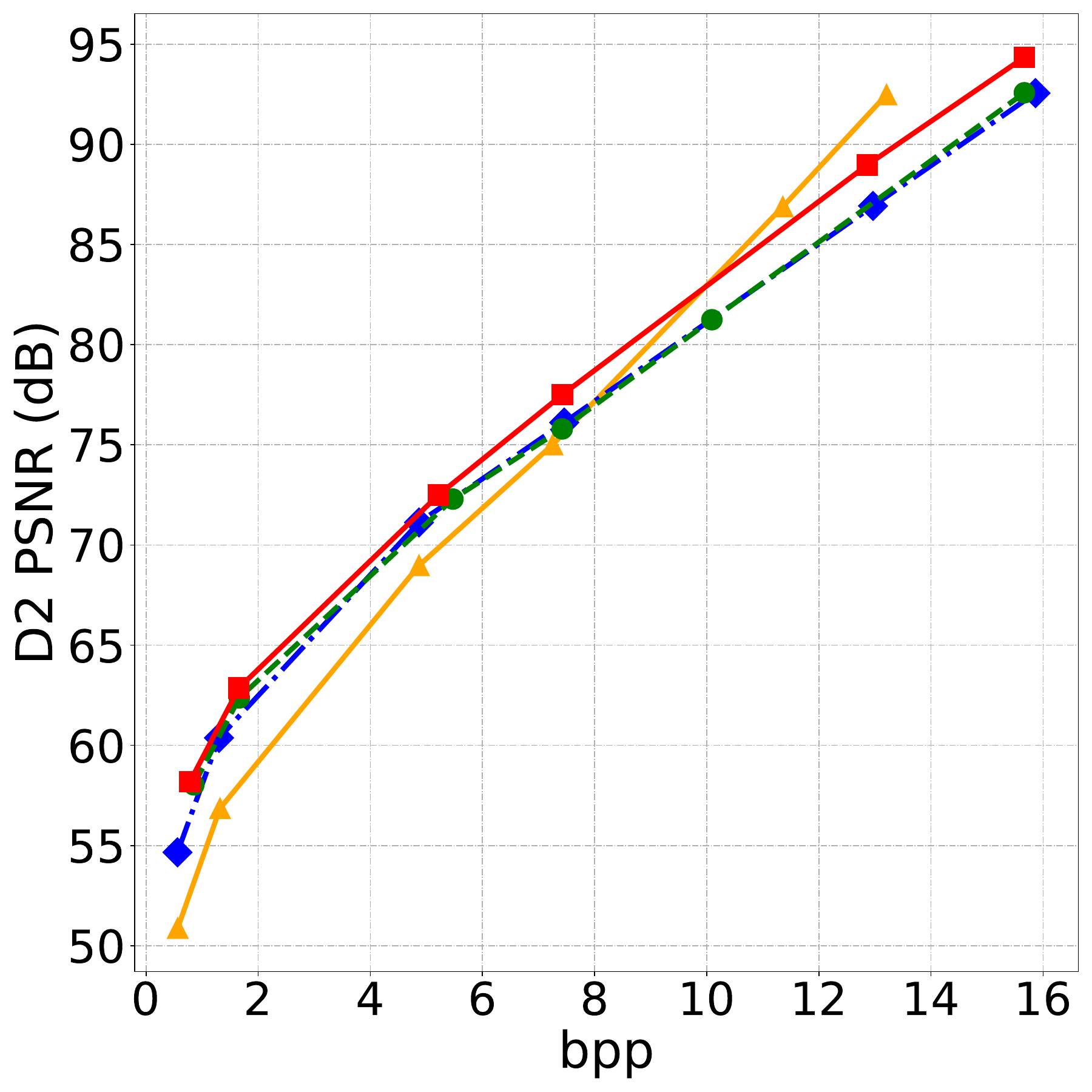}\label{fig:rd_lidar}}
  \vspace{-3pt}
  \caption{R-D performance on different point clouds measured in both D1-PSNR and D2-PSNR. Our proposed PIVOT-Net provides competitive performance among all the methods.}
  \label{fig:rd}
  \vspace{-2pt}
\end{figure*}

\begin{table*}[htbp]
  \centering\scriptsize
  \caption{BD-Rate (in \%) and BD-PSNR (in dB) gains against G-PCC octree (lossy) on different categories of point clouds.}\vspace{-8pt}
    \begin{tabular}{c||cc|cc||cc|cc||cc|cc||cc|cc}
    \hline
    Category & \multicolumn{4}{c||}{Solid surface point clouds} & \multicolumn{4}{c||}{Dense surface point clouds} & \multicolumn{4}{c||}{Sparse surface point clouds} & \multicolumn{4}{c}{LiDAR point clouds}\\
    \hline
    \multirow{2}[4]{*}{Metric} & \multicolumn{2}{c|}{BD-Rate $\downarrow$} & \multicolumn{2}{c||}{BD-PSNR $\uparrow$} & \multicolumn{2}{c|}{BD-Rate $\downarrow$} & \multicolumn{2}{c||}{BD-PSNR $\uparrow$} & \multicolumn{2}{c|}{BD-Rate $\downarrow$} & \multicolumn{2}{c||}{BD-PSNR $\uparrow$} & \multicolumn{2}{c|}{BD-Rate $\downarrow$} & \multicolumn{2}{c}{BD-PSNR $\uparrow$} \\
\cline{2-17}          & D1    & \multicolumn{1}{c|}{D2} & D1    & D2    & \multicolumn{1}{c}{D1} & \multicolumn{1}{c|}{D2} & \multicolumn{1}{c}{D1} & \multicolumn{1}{c||}{D2} & D1    & D2    & D1    & D2    & D1    & D2    & D1    & D2 \\
    \hline
    \hline
    GRASP-Net & -88.80 & -75.74 & 9.30  & 7.20  & -41.08 & -54.86 & 2.86  & 3.70  & -10.19 & -9.97 & 0.86  & 1.01  & -5.56 & -9.64 & 1.31  & 2.08 \bigstrut[t]\\
    SparsePCGC & -91.10 & -81.98 & \textbf{10.99} & \textbf{9.01} & -52.23 & \textbf{-68.10} & 4.26  & \textbf{6.11}  & 14.53 & 6.08  & -1.48 & -0.80 & -5.40 & -11.47 & 1.00  & 2.04 \\
    PIVOT-Net & \textbf{-92.34} & \textbf{-83.22} & 10.88 & 8.97  & \textbf{-58.42} & -66.61 & \textbf{4.37} & 5.20  & \textbf{-17.01} & \textbf{-21.21} & \textbf{1.40} & \textbf{2.30} & \textbf{-11.85} & \textbf{-17.19} & \textbf{2.19} & \textbf{3.10}\\
    \hline
    \end{tabular}%
  \label{tab:comp}%
  \vspace{-13pt}
\end{table*}%

We achieve different rate points by adjusting the interval $[n_1, n]$ (Fig.~\ref{fig:tvvp}) and train one model per rate point, where shorter interval $n-n_1$ means a deeper octree (or smaller block size) and leads to a larger rate point.
For surface point clouds, we sample $5$ sizes of $n-n_1$ in $[1, 4]$ for $5$ rate points.
Within $[n_1, n]$, we configure the sub-interval of point-based processing $[n_2, n]$ to have a size from $0$ to $4$ according to the surface point cloud categories.
Note that $n-n_2=0$ implies the point-based processing is turned off and the whole range $[n_1, n]$ is consumed by voxel-domain processing.
For the sparse LiDAR point clouds, we let $n-n_1$ ranges within $[2, 9]$ and pick $6$ rate points covering a typical operation range.
We also turn off the voxel-domain processing due to the highly sparse nature of LiDAR sweeps, \ie, $n_1=n_2$.
In all cases, we let $n_1-n'_1=1$, meaning that the feature map is 2$\times$ downsampled/upsampled for compression.

We apply the Adam optimizer~\cite{kingma2014adam} to train the PIVOT-Net for $50$ epochs with a learning rate $8\times10^{-4}$.
The batch size is $8$ when training on ModelNet40 and is $2$ when training on LiDAR data.
Please refer to the supplementary material for more details.

\textbf{Benchmarking}:
State-of-the-art lossy PCC methods are used for comparisons: (i) G-PCC octree (lossy), which is a non-learning-based method standardized by MPEG~\cite{graziosi2020overview}; (ii) GRASP-Net~\cite{pang2022grasp} and (iii) SparsePCGC~\cite{wang2022sparse} which are learning-based methods.
For fair comparisons, the octree coders in GRASP-Net, SparsePCGC, and our PIVOT-Net are all aligned as G-PCC octree (lossless).
The code/results of SparsePCGC and GRASP-Net are provided by their authors, where the SparsePCGC we compared is a slightly improved version~\cite{wang2022sparse2} over the original work~\cite{wang2022sparse}.

For evaluation, bitrate is measured by \emph{bits per input point} (bpp) while geometry distortion is measured by the peak signal-to-noise ratio (PSNR) based on \emph{point-to-point distance} (D1) and \emph{point-to-plane distance} (D2)~\cite{tian2017geometric}.
Following the convention in MPEG \cite{ctcgpcc}, we evaluate BD-Rate (in \%, the more negative it is, the more bitrate savings) and BD-PSNR (in dB, higher means better) by comparing an R-D curve with the one achieved by G-PCC octree (lossy), \ie, G-PCC octree (lossy) serves as an \emph{anchor}.

\def\vis_den{0.34\columnwidth}
\begin{figure*}[htbp]
  \centering \scriptsize
  {
  \includegraphics[width=1.55\columnwidth]{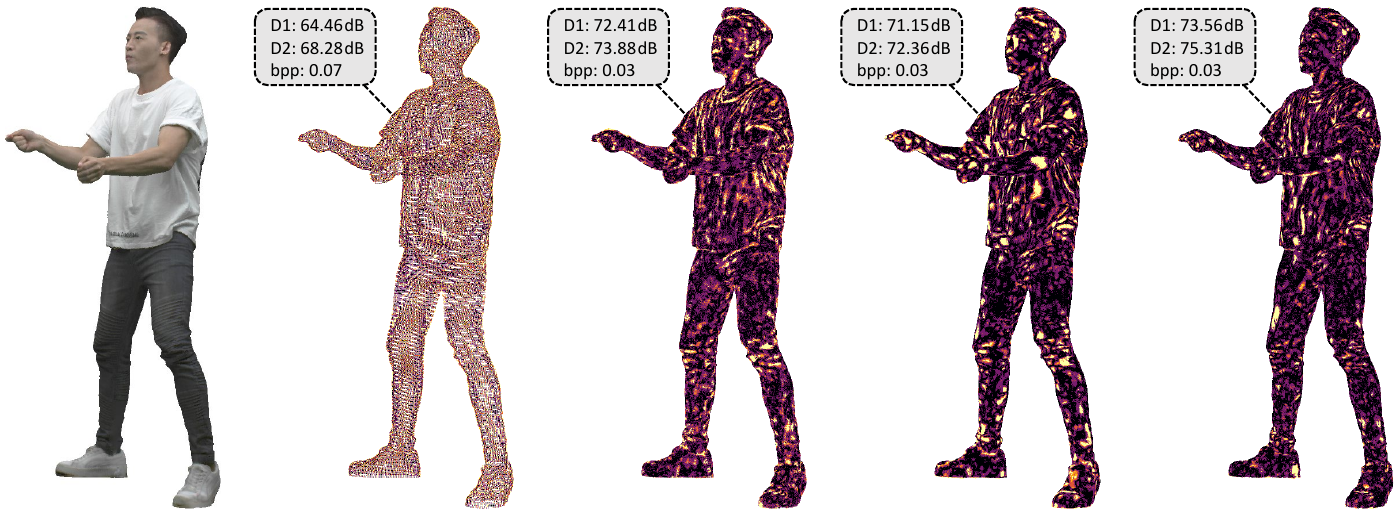}\hspace{15pt}\vspace{5pt}
  \includegraphics[height=130pt]{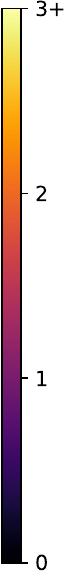}\\ \vspace{-15pt}
  \hspace{-22pt}
  \subfloat[Ground-truth]{\includegraphics[width=\vis_den]{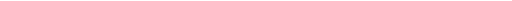}}
  \subfloat[G-PCC]{\includegraphics[width=\vis_den]{figures/place_holder.png}\label{fig:dense_gpcc}}
  \subfloat[GRASP-Net]{\includegraphics[width=\vis_den]{figures/place_holder.png}\label{fig:dense_spcgc}}
  \subfloat[SparsePCGC]{\includegraphics[width=\vis_den]{figures/place_holder.png}\label{fig:dense_skip}}
  \subfloat[PIVOT-Net]{\includegraphics[width=\vis_den]{figures/place_holder.png}\label{fig:dense_grasp}}
  }
  \vspace{-8pt}
  \caption{Visual comparisons of ``dancer\_vox11\_00000001'' (solid, 11-bit). The decoded point clouds are colored by the D1 errors.}
  \label{fig:vis_dense}
\end{figure*}

\def\vis_spa{0.36\columnwidth}
\begin{figure*}
  \centering \scriptsize
  \subfloat{\includegraphics[width=\vis_spa]{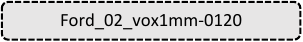}}\hspace{3pt}
  \subfloat{\includegraphics[width=\vis_spa]{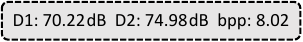}}\hspace{3pt}
  \subfloat{\includegraphics[width=\vis_spa]{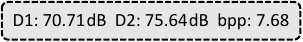}}\hspace{3pt}
  \subfloat{\includegraphics[width=\vis_spa]{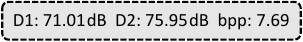}}\hspace{3pt}
  \subfloat{\includegraphics[width=\vis_spa]{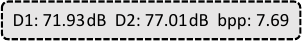}}
  \subfloat{\includegraphics[width=11pt]{figures/place_holder.png}}\\
  \setcounter{subfigure}{0}
  \subfloat[Ground-truth]{\includegraphics[width=\vis_spa]{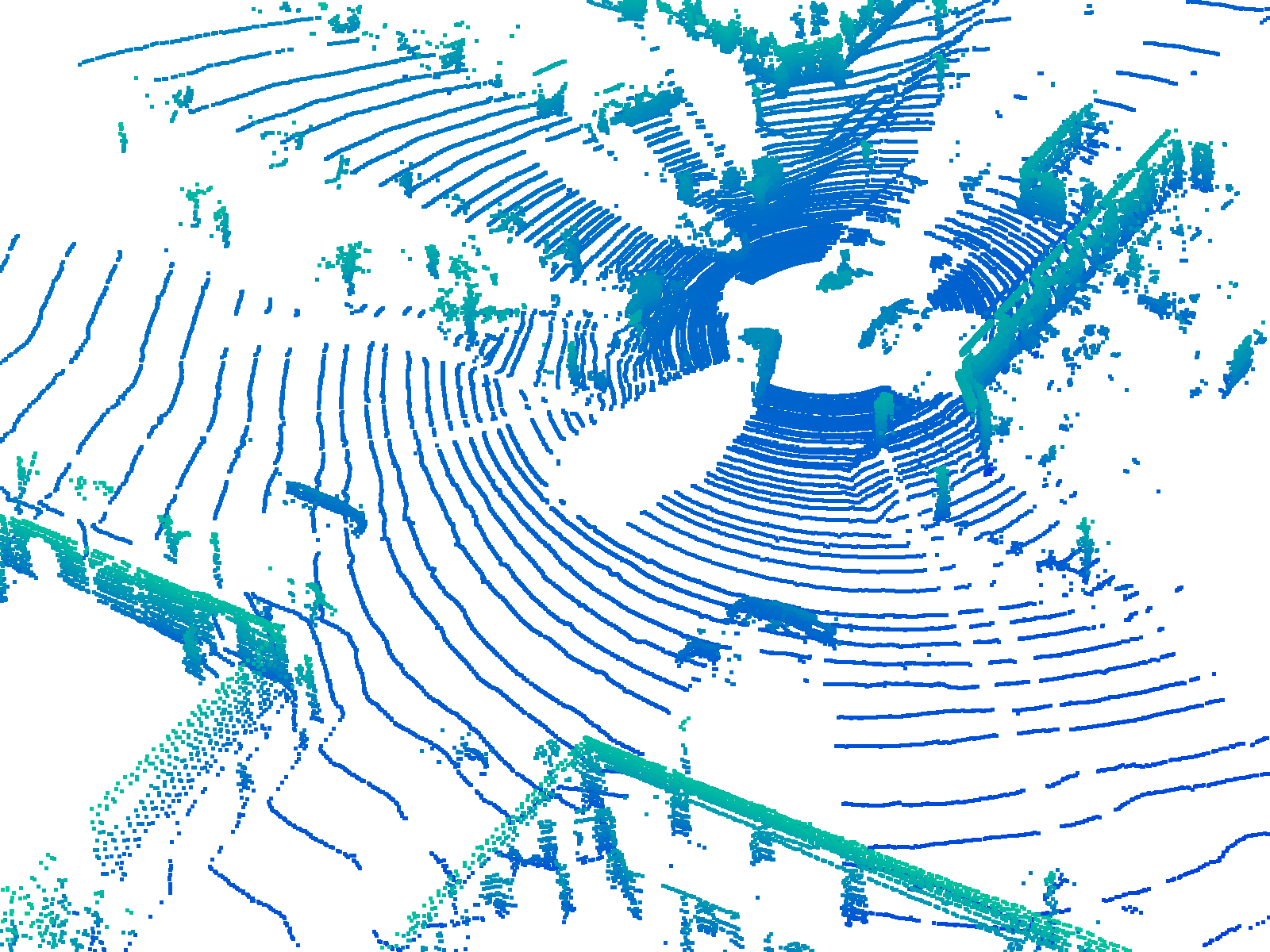}}\hspace{3pt}
  \subfloat[G-PCC]{\includegraphics[width=\vis_spa]{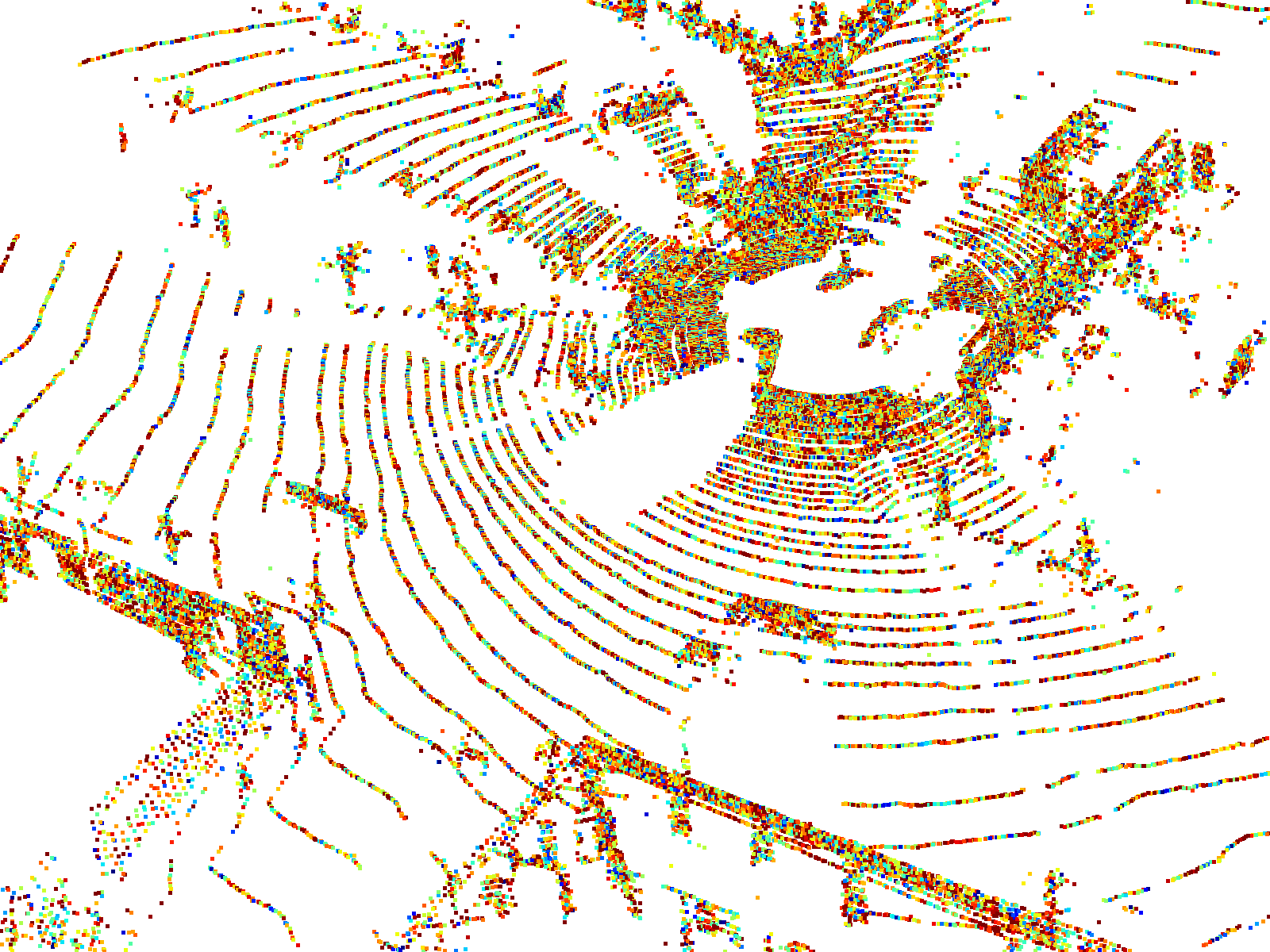}\label{fig:sparse_gpcc}}\hspace{3pt}
  \subfloat[GRASP-Net]{\includegraphics[width=\vis_spa]{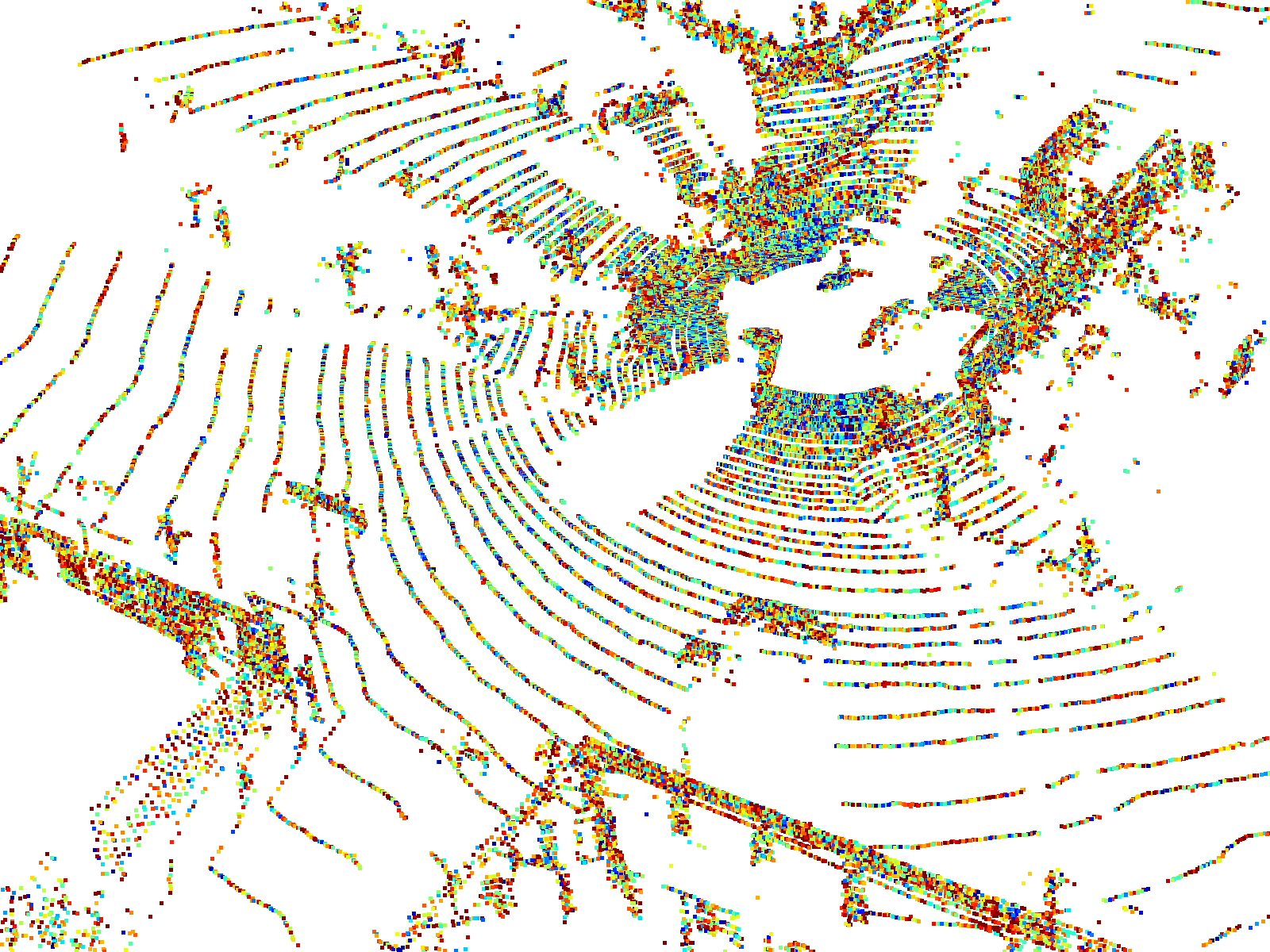}\label{fig:sparse_grasp}}\hspace{3pt}
  \subfloat[SparsePCGC]{\includegraphics[width=\vis_spa]{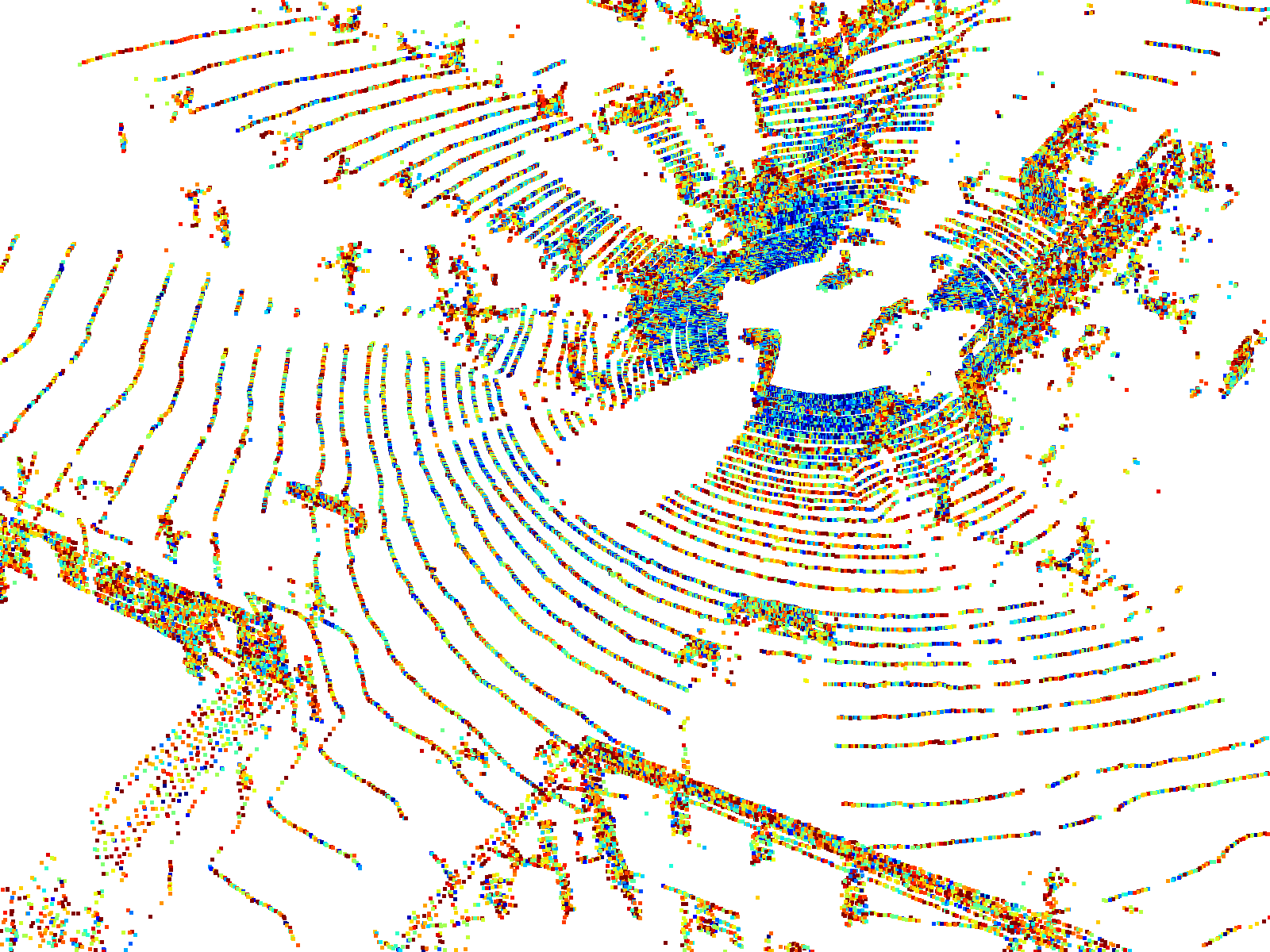}\label{fig:sparse_spcgc}}\hspace{3pt}
  \subfloat[PIVOT-Net]{\includegraphics[width=\vis_spa]{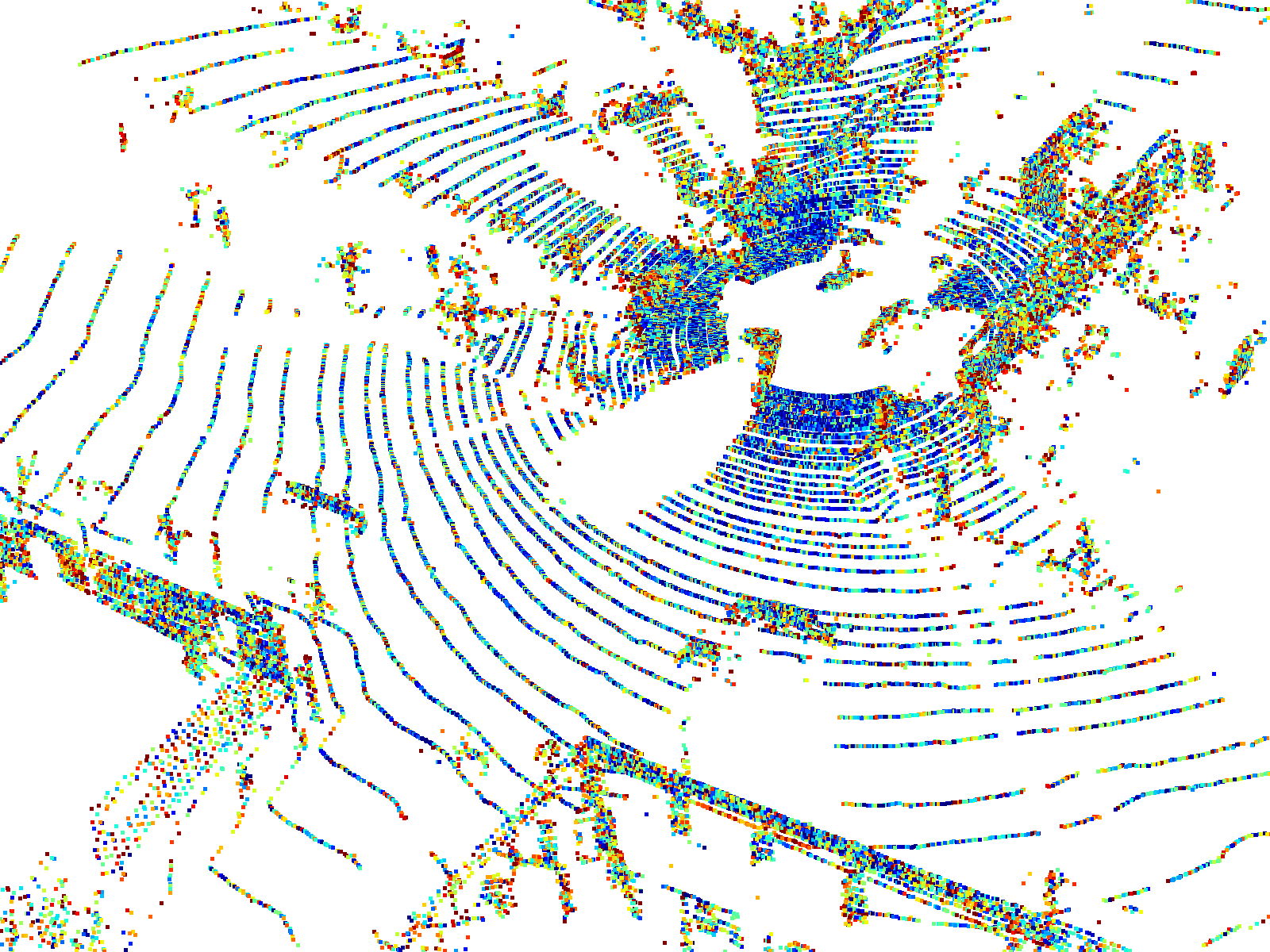}\label{fig:sparse_pivot}}
  \hspace{4pt}\subfloat{{\includegraphics[width=11pt]{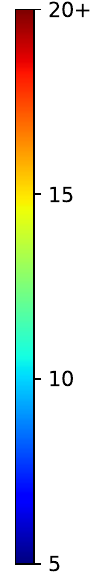}}}\\
  \vspace{-8pt}
  \caption{Visual comparisons of LiDAR point cloud ``ford\_02\_vox1mm-0120'' (18-bit). Decoded point clouds are colored by D1 errors.}
  \label{fig:vis_sparse}
  \vspace{-13pt}
\end{figure*}

\vspace{-2pt}
\subsection{Performance Comparisons}
\vspace{-3pt}
The average R-D curves of the different methods on representative point clouds are plotted in Fig.~\ref{fig:rd}.
We also compute the BD-Rate and BD-PSNR gains against the G-PCC octree (lossy) anchor.
The results are provided Table~\ref{tab:comp}.

Overall, we see that our PIVOT-Net achieves the best R-D trade-off among the competing approaches.
It performs on par with SparsePCGC~\cite{wang2022sparse} on solid and dense surface point clouds while outperforming SparsePCGC significantly on sparse surface point clouds and LiDAR point clouds.
That is because we account for the sparse nature of these point clouds with point-based processing (the $[n_2, n]$ interval of Fig.~\ref{fig:tvvp}) while SparsePCGC does not.
Compared to GRASP-Net~\cite{pang2022grasp}, we achieve clear advantages on the solid and dense surface point clouds because of the consumption of the middle-range bits in the voxel domain (the $[n_1, n_2]$ interval of Fig.~\ref{fig:tvvp}) while GRASP-Net ignores that.

Fig.~\ref{fig:vis_dense} visualizes the $11$-bit solid surface point cloud ``dancer\_vox11\_00000001'' decoded by different methods while Fig.~\ref{fig:vis_sparse} visualizes the decoded results of the LiDAR sweep ``ford\_02\_vox1mm-0120''.
The decoded point clouds in these figures are colorized by their D1 errors.
Compared with the other methods, our PIVOT-Net provides higher-quality reconstructions at smaller or similar bit-rates.
Notably, it makes fewer errors in intricate details such as the corners in Fig.~\ref{fig:vis_dense} and the line patterns in Fig.~\ref{fig:vis_sparse}.

\vspace{-2pt}
\subsection{Ablation Study}
\vspace{-3pt}
We study how the coding performance of the PIVOT-Net is benefited by (i)~the voxel processing of the middle-range bits and (ii)~the transformer.
On solid surface point clouds, the performance of several variations of the PIVOT-Net are shown in Table~\,\ref{tab:abla}.
We start from a configuration where both (i) and (ii) are removed (\ie, GRASP-Net~\cite{pang2022grasp}).
Then we enable (i) (first without, then with the context-aware upsampling) which leads to around $3\%$ and $6\%$ BD-Rate gains in terms of D1 and D2, respectively.
Next, we enable (ii) (with the voxel transformer~\cite{mao2021voxel}, then with the Enhanced Voxel Transformer), bringing an extra $1\%$ BD-Rate saving.
Thus, we verify the importance of (ii) for PIVOT-Net.

\begin{table}[t]
  \centering\scriptsize
  \caption{BD-Rate (in \%) and BD-PSNR (in dB) gains against G-PCC octree (lossy) on different variations of the PIVOT-Net.}
  \vspace{-4pt}
    \begin{tabular}{cc|cc||cc|cc}
    \hline
    \multicolumn{2}{c|}{Voxel for mid. bits} & \multicolumn{2}{c||}{Transformer} & \multicolumn{2}{c|}{BD-Rate $\downarrow$} & \multicolumn{2}{c}{BD-PSNR $\uparrow$} \\
    \hline
    w/o ctx.    & w/ ctx. & VT & Enh. VT & D1    & D2    & D1    & D2 \\
    \hline
    \hline
    $\times$     & $\times$     & $\times$     & $\times$     & -88.80 & -75.74 & 9.30  & 7.20 \\
    \checkmark     & $\times$     & $\times$     & $\times$     & -91.31 & -81.46 & 10.52 & 8.70 \\
    $\times$     & \checkmark     & $\times$     & $\times$     & -91.72 & -82.01 & 10.64 & 8.80 \\
    $\times$     & \checkmark     & \checkmark     & $\times$     & -91.85 & -82.40 & 10.58 & 8.79 \\
    $\times$     & \checkmark     & $\times$     & \checkmark     & \textbf{-92.34} & \textbf{-83.22} & \textbf{10.88} & \textbf{8.97} \\
    \hline
    \end{tabular}%
  \label{tab:abla}%
\vspace{-6pt}
\end{table}%

We also adjust the intervals for solid (most dense) and LiDAR (most sparse) point clouds, showing the trade-offs in Fig.~\ref{fig:rd_interval}. In Fig.~\ref{fig:rd_interval}, each operation point in the R-D plot is denoted by $[c,m,f]$, where $c=n_1$, $m=n_2-n_1$, and $f=n-n_2$ bits are the coarsest-, middle-, and finest-bit intervals (Fig.~\ref{fig:tvvp}), respectively.
From Fig.~\ref{fig:solid}, $\red{\blacktriangledown}$ operation points (with $m>0$, \ie, voxel processing for middle bits) are preferable for solid point clouds, where the $\red{\blacktriangledown}$ points delineate the best R-D curve.
But for sparse point clouds (Fig.~\ref{fig:lidar}), point-based coding ($\blue{\blacktriangle}$) is preferred.

\def\rd_hght{55pt}
\begin{figure}[t]
  \centering \scriptsize
  \subfloat[``dancer\_vox11'' (solid, 11-bit)]{\includegraphics[height=\rd_hght]{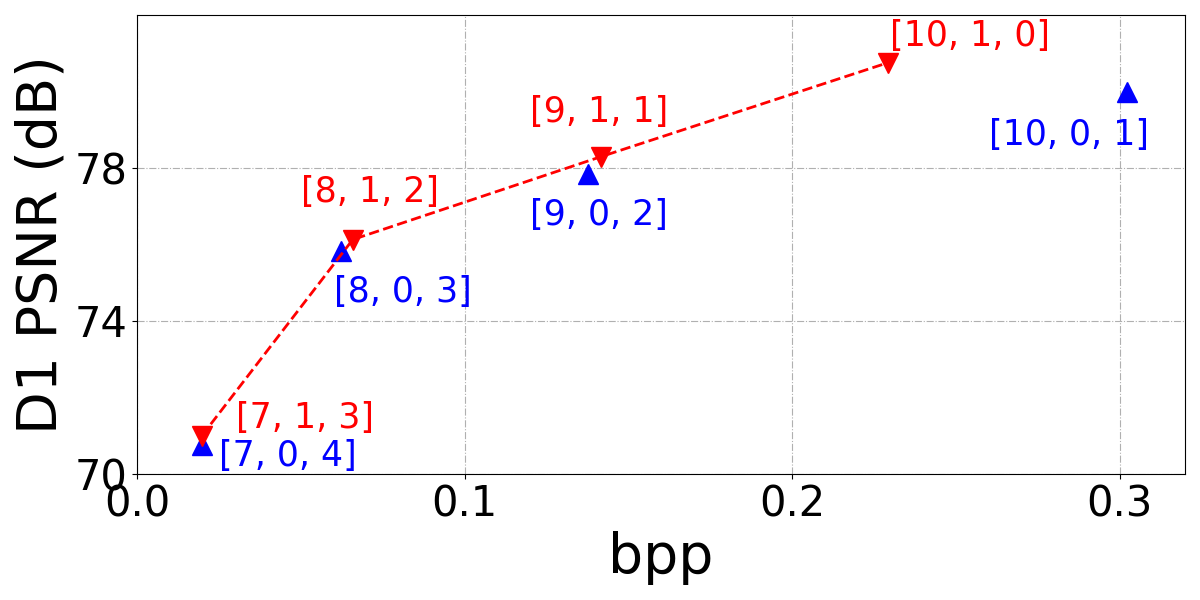}\label{fig:solid}}\hspace{3pt}
  \subfloat[``ford\_02\_vox1mm'' (18-bit)]{\includegraphics[height=\rd_hght]{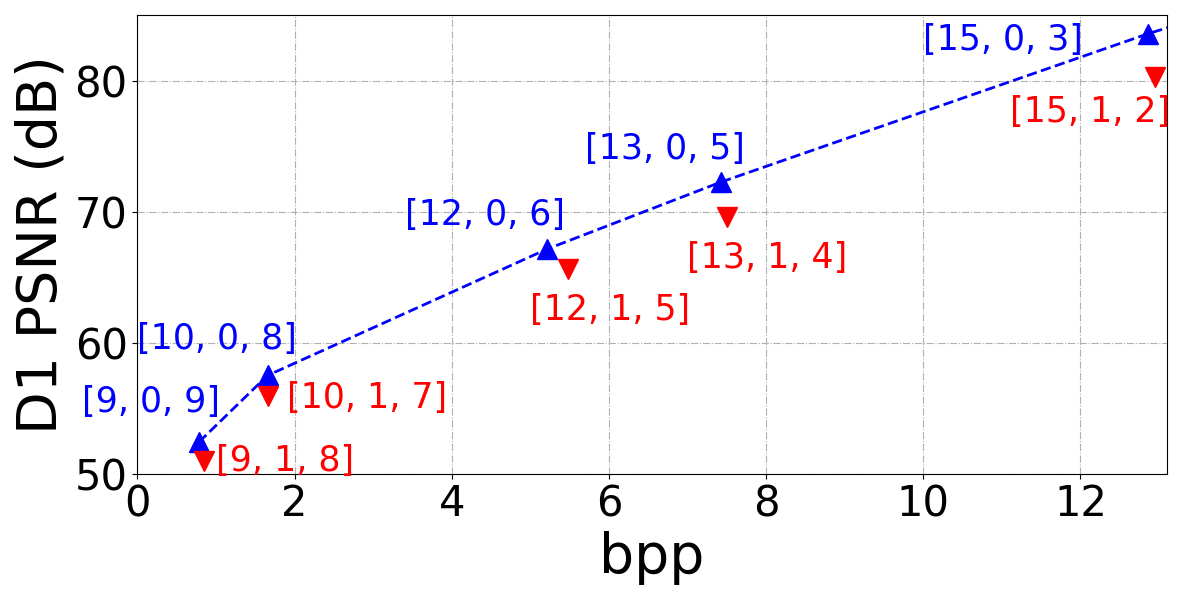}\label{fig:lidar}}
  \vspace{-8pt}
  \caption{Changing the interval lengths in PIVOT-Net.}
  \label{fig:rd_interval}
  \vspace{-17pt}
\end{figure}

\subsection{Complexity}
PIVOT-Net has a reasonable computational cost.
For instance, on the $3000$ test LiDAR frames, to encode/decode a frame at around $8$~bpp (the rate point of the examples in Fig.~\ref{fig:vis_sparse}), PIVOT-Net takes about $1.8$~sec/$2.0$~sec.
On the same platform, G-PCC takes $3.8$~sec/$1.8$~sec, GRASP-Net takes $1.6$~sec/$1.7$~sec and SparsePCGC takes $2.1$~sec/$11.5$~sec.
Note that running time is measured with the \texttt{time.monotonic()} function in Python.

PIVOT-Net also has a small model size.
When compressing different point cloud categories, neural network models of different sizes are applied.
On average, the models of PIVOT-Net have $482$\,K parameters, being slightly larger than those of the GRASP-Net ($430$\,K).
However, both of them are much smaller than the model size of SparsePCGC, which on average has $3363$\,K parameters.

%% file: sec/6_conclusion.tex
We propose PIVOT-Net, a heterogeneous PCC framework unifying three point cloud representations---the point-based, voxel-based, and tree-based representations---to altogether digest the least-significant, the middle-range, and the most-significant bits.
The voxel-domain processing is augmented with a proposed context-aware upsampling procedure for decoding and an Enhanced Voxel Transformer for feature aggregation.
The state-of-the-art performance of PIVOT-Net is confirmed on a wide spectrum of point clouds.
A future research direction is to optimize the allocation of the bits assigned to the three representations.